\documentclass[10pt,journal,cspaper,compsoc]{IEEEtran}
\usepackage{graphicx}
\usepackage{amsmath,amssymb}
\usepackage[usenames, dvipsnames]{color}
\usepackage{times}
\usepackage{multirow}
\usepackage[caption=false]{subfig}
\usepackage{ragged2e}
\usepackage{multirow,array}
\usepackage{cite}
\usepackage{rotating}
\usepackage{multicol,multirow}
\usepackage{eqparbox}
\usepackage{overpic}
\usepackage{pifont}% http://ctan.org/pkg/pifont
\usepackage[colorlinks,bookmarksnumbered,bookmarksopen,citecolor=black]{hyperref}
\usepackage{silence}
\usepackage{xspace}
\def\eg{\textit{e.g.}}
\def\ie{\textit{i.e.}}

\def\etal{\textit{et al. }}

\newcommand{\dimny}{\mathcal{M}\xspace}
\newcommand{\distribution}{\mathcal{D}\xspace}

\newcommand{\dimH}{\ensuremath{H}}
\newcommand{\dimW}{\ensuremath{W}}
\newcommand{\dimC}{\ensuremath{C}}
\newcommand{\nStage}{\mathcal{Q}\xspace}

\newcommand{\mypartitletwo}[1]{\vspace*{-.5ex}~\\{\noindent \underline{\bf #1}}}

\newcommand{\dimn}{\ensuremath{M}}

\newcommand{\mapping}{\ensuremath{G}\xspace}
\newcommand{\mappingfunc}{\mathcal{G}\xspace}
\newcommand{\loss}{\mathcal{L}\xspace}

\newcommand{\params}{\ensuremath{\theta}\xspace}
\newcommand{\data}{\ensuremath{X}\xspace}

\newcommand{\featSpace}{\ensuremath{\mathrm{\cal X}}\xspace}
\newcommand{\lSpace}{\ensuremath{\mathrm{\cal Y}}\xspace}

\newcommand{\nsamples}{\ensuremath{N}\xspace}

\newcommand{\nTrees}{\ensuremath{K}\xspace}

\newcommand{\Ymat}{\ensuremath{\mathbf{Y}}\xspace}

\newcommand{\samp}{\ensuremath{\mathbf{x}}\xspace}
\newcommand{\laby}{\ensuremath{\mathbf{y}}\xspace}

\newcommand{\RR}{I\!\!R}

\newcommand{\mymodel}{DNCL}

\def\etal{{\em et al.~}}

\newcommand{\cmark}{\ding{51}}%
\newcommand{\xmark}{\ding{55}}%

\def\sArt{{state-of-the-art~}}

\DeclareGraphicsExtensions{.pdf,.jpg,.png}
\newcommand{\figref}[1]{Fig.~\ref{#1}}
\newcommand{\tabref}[1]{Table~\ref{#1}}
\newcommand{\secref}[1]{Section~\ref{#1}}

\newcommand{\equref}[1]{Eqn. (\ref{#1})}
\newcommand{\myPara}[1]{\vspace{.25in}\noindent\textbf{#1}~~}

\usepackage{silence}
\graphicspath{{./image/}{./image/author/}}

\newcommand{\zl}[1]{{{#1}}}

\newcommand{\RowsT}[1]{\multirow{2}{*}{#1}}
\newcommand{\ColsT}[2]{\multicolumn{2}{c#1}{#2}}

\begin{document}

\title{Robust Regression via Deep Negative Correlation Learning}

\author{Le~Zhang,
  Zenglin~Shi, Ming-Ming Cheng, Yun~Liu, Jia-Wang Bian,\\
  Joey Tianyi Zhou, Guoyan Zheng and Zeng Zeng % <-this % stops a space
 
  \IEEEcompsocitemizethanks{ \IEEEcompsocthanksitem  The first two authors are the joint first author and Ming-Ming Cheng is the corresponding author.
  \IEEEcompsocthanksitem 
    Le Zhang, Joey Tianyi Zhou and Zeng Zeng are with the Agency for Science, 
    Technology and Research (A*STAR), Singapore.
  \IEEEcompsocthanksitem Zenglin Shi is with the University of Amsterdam.
  \IEEEcompsocthanksitem Guoyan Zheng is with the School of Biomedical Engineering,  Shanghai JiaoTong University, China.
  \IEEEcompsocthanksitem Ming-Ming Cheng, Yun Liu are with the Nankai University, China.
  \IEEEcompsocthanksitem Jia-Wang Bian is with the School of Computer Science, The University of Adelaide, Adelaide, Australia}% <-this % stops a space
  \thanks{Manuscript received April 19, 2005; revised August 26, 2015.}
}

\markboth{IEEE Transactions on Pattern Analysis and Machine Intelligence}%
{Zhang \MakeLowercase{\textit{et al.}}: Robust Regression via Deep Negative Correlation Learning}

\IEEEtitleabstractindextext{%
\begin{abstract}
\justifying
Nonlinear regression has been extensively employed in many computer vision
problems (e.g., crowd counting, age estimation, affective computing).
Under the umbrella of deep learning, two common solutions exist
i) transforming nonlinear regression to a robust loss function
which is jointly optimizable with the deep convolutional network, and
ii) utilizing ensemble of deep networks.
Although some improved performance is achieved,
the former may be lacking due to the intrinsic limitation of
choosing a single hypothesis and the latter usually suffers from
much larger computational complexity.
To cope with those issues,
we propose to regress via an efficient ``divide and conquer" manner.
The core of our approach is the generalization of negative correlation
learning that has been shown,
both theoretically and empirically,
to work well for non-deep regression problems.
Without extra parameters,
the proposed method controls the bias-variance-covariance trade-off
systematically and usually yields a deep regression ensemble
where each base model is both ``accurate" and ``diversified."
Moreover, we show that each sub-problem in the proposed method has
less Rademacher Complexity and thus is easier to optimize.
Extensive experiments on several diverse and challenging tasks
including crowd counting, personality analysis, age estimation,
and image super-resolution demonstrate the superiority over
challenging baselines as well as the versatility of the proposed method.

\end{abstract}

}

\maketitle

\IEEEdisplaynontitleabstractindextext

\IEEEpeerreviewmaketitle

% ========================== Introduction ============================
\section{Introduction}\label{sec:introduction}

\IEEEPARstart{W}{e} address regression problems which aim at
analyzing the relationship between dependent variables (targets)
and independent variables (inputs).
Regression has been applied to a variety of computer vision problems
including crowd counting~\cite{shi2018crowd},
age estimation~\cite{huo2016deep},
affective computing~\cite{ponce2016chalearn},
image super-resolution~\cite{tai2017image},
visual tracking~\cite{zhang2017robust} and so on.
Pioneering works within this area typically learn a mapping function
from hand-crafted features (\eg, Histogram of Oriented Gradients (HoG),
Scale-Invariant Feature Transform (SIFT)) to the desired
output (\eg, ages, affective scores, density maps and so on).

Recently, transforming a regression problem to an optimizable
robust loss function
%(\eg, L2, L1, Huber~\cite{girshick2015fast}, and
%Tukey’s biweight loss~\cite{belagiannis2015robust})
jointly trained with deep Convolutional Neural Network (CNN)
has been reported to be successful to some extent.
Most of the existing deep learning based regression approaches optimize
the L2 loss function together with a regularization term,
where the goal is to minimize the mean square error between
the network prediction and the ground-truth.
However, it is well known that the mean square error is
sensitive to outliers,
which are essentially the samples that lie at an abnormal distance
from other training samples in the objective space.
In this case, samples that are rarely encountered in the training data
may have a disproportionally high weight and consequently
influence the training procedure by reducing the generalization ability.
To this end, Ross Girshick~\cite{girshick2015fast} introduced a SmoothL1
loss for bounding box regression.
As a special case of Huber loss~\cite{huber1964robust},
the SmoothL1 loss combines the concept of L2 loss and L1 loss.
It behaves as an L1 loss when the absolute value of the error is high
and switches back to L2 loss when the absolute value of the error
is close to zero.
Besides, Belagiannis~\etal\cite{belagiannis2015robust} propose
a deep regression network that achieves robustness to outliers by
minimizing Tukey’s biweight function
\cite{huber2011robust,black1996unification} .

While tremendous progress has been achieved later by
employing robust statistical estimations together with
specially-designed network architecture to explicitly address outliers,
they may fail to generalize well in practice.
As studied in~\cite{dietterich2000ensemble},
a single model was lacking due to the statistical,
computational and representational limitations.
To this end, a  great deal of research has gone into designing
multiple regression systems
\cite{dietterich2000ensemble,breiman2001random,ren2016ensemble,qiu2017oblique}.
However, existing methods for ensemble CNNs
\cite{walach2016learning,han2016incremental,cortes2014deep}
typically trained multiple CNNs,
which usually led to much larger computational complexity
and hardware consumption.
Thus, these ensemble CNNs are rarely used in practical systems.
%They were usually very slow in terms of both training and inference,
%which naturally limited their applicability for real-life applications.

In this paper, we propose a Deep Negative Correlation Learning (DNCL)
approach which learns a pool of diversified regressors in
a ``divide and conquer" manner.
Each regressor is jointly optimized with CNNs by an amended cost function,
which penalizes correlations with others.
Our approach inherits the advantage of traditional \emph{Negative Correlation
Learning (NCL)}~\cite{liu2000evolutionary,brown2005managing} approaches,
that systematically controls the trade-offs among the
\emph{bias-variance-covariance} in the ensemble.
Firstly, by dividing the task into multiple ``negatively-correlated" sub-problems,
the proposed method shares the essence of ensemble learning and yield more robust
estimations than a single network~\cite{ren2016ensemble,brown2005managing}.
Secondly, thanks to the rich feature hierarchies in deep networks,
each sub-problem could be solved by a feature subset.
In this way, the proposed method  has a similar amount of parameters
with a single network and thus is much more efficient
than most existing deep ensemble learning
{\cite{walach2016learning,han2016incremental,cortes2014deep}}.
Simplicity and efficiency are central to our design,
the proposed methods are almost complementary to other advanced strategies
for individual regression tasks.

A preliminary version of this work was presented in CVPR
2018~\cite{shi2018crowd},
which provides an application of DNCL for crowd counting.
%
% This paper adds to the initial version in significant ways.
% We provide more theoretical insights on the Rademacher complexity
% for the proposed method.
% We also extend the original work to deal with more regression based problems,
% which allows the use of state-of-the-art network structures that give
% an important boost to performance.
% More comprehensive literature review,
% considerable new analysis and intuitive explanations are added to the
% initial results.
This paper adds to the initial version in the following aspects:
\begin{itemize}
    \item We provide more theoretical insights on the Rademacher complexity.
    \item We extend the original work to deal with more regression based problems, which allows the use of state-of-the-art network structures that give an important boost to performance for the proposed method.
    \item More comprehensive literature review,
considerable new analysis and intuitive explanations are added to the
initial results.
\end{itemize}

% In summary, we make the following contributions:

% ========================== Related work ============================
\section{Related Work}

\subsection{Regression}
We first briefly introduce the commonly used loss function for
regression based deep learning computer vision tasks,
followed by summarizing the existing ensemble regression techniques.

\myPara{Deep Regression.} Recently, learning a mapping function to predict a set of
interdependent continuous values by deep networks is popular.
One example could be object detection where the target is
to regress the bounding box for precise localization~\cite{ren2015faster}.
Other examples include regressing the facial points in
facial landmark detection~\cite{sun2013deep} and
positions of the body in human pose estimation~\cite{toshev2014deeppose}.
The L2 loss function is a natural choice for solving such problems.
%In such typical problems, CNNs were trained just using an normal L2 loss function,
%without considering any further enhanced learning strategy.
Zhang \etal\cite{zhang2014facial} further utilized L2 regularization
to increase the robustness of network
for both landmark detection and attribute classification.
Similar strategies were also applied in object detection~\cite{wang2014deep}.

The commonly used L2 loss in regression problems may not generalize well
in the case of outliers because outliers can have a disproportionally high weight,
and consequently influence the training procedure by reducing the
generalization ability and increasing the convergence time.
To this end, a SmoothL1 loss~\cite{girshick2015fast} was reported to be
more robust than L2 loss when outliers are present in the dataset:
\begin{equation}
  \rm{Smooth_{L_1}}(\xi )=
  \left\{\begin{matrix}
    0.5\xi ^2  & if ~|\xi |<1 \\
    |\xi |-0.5 & otherwise
  \end{matrix} \right. ,
  \label{equ:smoothL1}
\end{equation}
where $\xi$ stands for the prediction error.
Motivated by the recent success in robust statistics~\cite{huber2011robust},
an M-estimator based~\cite{black1996unification} loss function,
called Tukey Loss~\cite{belagiannis2015robust},
was proposed for both human pose estimation and age estimation
\cite{belagiannis2015robust}.
More specifically,
\begin{equation}
  \rm{Tukey}(\hat{\xi})=
  \left\{\begin{matrix}
    \frac{c^2}{6}[1-(1-(\frac{\hat{\xi}}{c})^2)^3] & if~ |\hat{\xi}|\leq c \\
    \frac{c^2}{6} & otherwise
  \end{matrix}\right. ,
  \label{equ:tukey}
\end{equation}
where $c$ is a tuning parameter, and is commonly set to $4.6851$,
which gives approximately $95\%$ asymptotic efficiency as L2 minimization
on the standard normal distribution of residuals.
$\hat{\xi}$ is a scaled version of
the residual $\xi$ by computing the median absolute deviation by:
\begin{equation}
  \begin{aligned}
    \hat{\xi} &= \frac{\laby-\bar{\laby}}{1.4826\times MAD},\\
    \rm{MAD} &= \underset{i \in\{1,\cdot,\nsamples\}}{\rm{median}}(
      |%\begin{bmatrix}
        \xi_i-\underset{k \in\{1,\cdot,\nsamples\}}{median}(\xi_k)
      |%\end{bmatrix}
      ),
  \end{aligned}
\end{equation}
where $\laby$, $\bar{\laby}$ and $\nsamples$ stands for the ground-truth label,
predicted result and number of data samples, respectively.
In case of regressing multiple outputs, the MAD values are calculated independently.

Our proposed DNCL method could also be regarded as a loss function,
which is readily pluggable into existing %any
CNN architecture and amenable to training via backpropagation.
Without extra parameters, the proposed methods mimic ensemble learning and
have a better control of the trade-off between the intrinsic bias,
variance and co-variance.
We evaluate the proposed method on multiple challenging and diversified regression tasks.
When combined with the state-of-the-art network structure,
our method could give an important boost to the performance of existing loss functions
mentioned above.

\myPara{Ensemble Regression.} Ensemble methods are wildly regarded to be better than single model if the ensemble
is both ``accurate" and ``diversified"
\cite{dietterich2000ensemble,breiman2001random,ren2016ensemble}.
As studied in~\cite{dietterich2000ensemble},
a single model was less generalizable from the statistical,
computational and representational point of view.
To this end, a bunch of research has gone into designing multiple regression systems.
For instance, the accuracy and diversity in a typical decision tree ensemble
\cite{breiman2001random,barandiaran1998random,zhang2017robust,qiu2017oblique}
were guaranteed by allowing each decision tree grow to its maximum depth and
utilizing feature subspace, respectively.
Boosting~\cite{friedman2001greedy} generated a new regressor with an amended
loss function based on the loss of the existing ensemble models.

Motivated by the success of ensemble methods,
several deep regression ensemble methods were proposed as well.
However, existing methods for training CNN ensemble
\cite{walach2016learning, cortes2014deep}
usually generated multiple CNNs separately.
In this case, the resulting system usually yielded a much larger
computational complexity compared with single models and
thus were usually very slow in terms of both training and inference,
which naturally limited their applicability for resource-constrained scenarios.

One of the exceptions could be the Deep Regression Forest (DRF)~\cite{Shen_2018_CVPR}
which reformulated the split nodes as a fully connected layer of a CNN and
learnt the parameter of CNN and tree nodes jointly by an alternating strategy.
Firstly, by fixing the leaf nodes, the internal nodes of trees as well as the CNN
were optimized by back-propagation.
After that,  both the CNN and the internal nodes were frozen and the leaf nodes
were learned by iterating a step-size free and fast converging update rule
derived from Variational Bounding.
We show the proposed method could also be combined with the concept of DRF.
The resulting system is much simpler to learn and yield a significant
improvement enhancement,
as elaborated in \secref{sec:age_exp}.

\subsection{Applications}

The proposed method is generic and could be applied to a wide range
of regression tasks.
It mimics ensemble learning without extra parameters and helps to learn more
generalizable features through a better control of the trade-off between
the intrinsic bias, variance and co-variance.
We evaluated it on multiple challenging and diversified regression tasks
including crowd counting, age estimation, personality analysis and
image super-resolution.
Simplicity is central to our design and the strategies adopted in the proposed method
are complementary to many other specially-designed techniques for each task.
When combined with state-of-the-art network structure for each task,
our proposed method is able to yield an important boost to the baseline methods.
Below we provide a detailed review on the recent advances in each task.

% Detecting people individually~\cite{zhao2008segmentation} seems to be a straightforward solution for crowd counting but suffers from severe defects: detectors are prone to fail when people are in close proximity and there is no way to recover. Furthermore, high computational complexities in detection based approaches also limit their applicability for real-time applications. Counting by regression, on the other hand, learns to predict pedestrians' number through a regression function with some visual descriptors such as texture features, edge features \cite{ryan2015evaluation} or learned representations~\cite{zhang2015cross,zhang2016single}.

\myPara{Crowd Counting.}
Counting by regression is perceived as the \sArt at present.
The regression-based methods have been widely studied and reported to be
computationally feasible with modern hardware, robust with parameters
and accurate across various challenging scenarios.
A deep CNN~\cite{zhang2015cross} was trained alternatively with two related
learning objectives, crowd density classification and crowd counting.
However, it relied heavily on a switchable learning approach and was not
clear how these two objective functions can alternatively assist each other.
Wang \etal cite{wang2015deep} proposed to directly regress the total people number
by adopting AlexNet~\cite{krizhevsky2012imagenet},
which has now been found to be worse than the methods regressing density map.
This observation suggests that reasoning with rich spatial layout
information from convolutional feature maps is necessary.
Boominathan \etal \cite{boominathan2016crowdnet} proposed a framework consisting of
both deep and shallow networks for crowd counting.
It was reported to be more robust with scale variations,
which have been addressed explicitly by other studies
\cite{shi2018multiscale,zhang2016single,onoro2016towards} as well.
Switching CNN was introduced in~\cite{sam2017switching},
where patches from a grid within a crowd scene were relayed to
independent CNN regressors based on crowd count prediction
quality of the CNN established during training.
\zl{Arteta \etal \cite{arteta2016counting} augmented and interleave
density estimation with foreground-background segmentation and explicit local uncertainty estimation under a new deep multi-task architecture. Noroozi \etal \cite{noroozi2017representation} used
counting as a pretext task to train a neural network with a
contrastive loss and showed improved results on transfer learning benchmarks.}

% Almost all the aforementioned approaches work well for their adopted CNN structures. Generating them to a much deeper network structure like~\cite{simonyan2014very} to further boost the discriminative ability of the learned representations for crowd counting is not straight-forward due to limited training data. Introducing large-scale datasets for crowd counting may partially alleviate the problem. However, manual labeling  is costly, time-consuming and error-prone. It can also raise privacy concerns. It is often impractical as there may exist several thousands of people within a single image in dense crowd scenarios. This motivates us to study the problem of training deep CNNs on existing crowd counting datasets with less risk of over-fitting. To address this, we draw inspirations from NCL~\cite{liu1999ensemble,brown2005managing} and extend it to deep learning.  With no extra learning parameter, it learns an ensemble of accurate and diversified regressors for crowd counting whose prediction errors may cancel out each other.

\myPara{Personality Analysis.}
Recent personality-related work with visual cues attempted to identify
personality from body movement~\cite{lepri2012connecting},
facial expression change~\cite{biel2012facetube,sanchez2013inferring},
combining acoustic cues~\cite{abadi2015inference},
eye gaze~\cite{batrinca2011please}, and so on.
In addition, recognizing personality traits using deep learning on images or videos
has also been extensively studied.
`ChaLearn 2016 Apparent Personality Analysis competition' \cite{ponce2016chalearn}
provided an excellent platform,
where researchers could assess their deep models on a large annotated
big-five personality traits dataset.
Instead of classifying pre-defined personality categories,
common practices use a finer-grained representation,
in which personalities are distributed in a five-dimensional space
spanned by the dimensions of \emph{Extraversion,
Agreeableness, Conscientiousness, Neuroticism
and Openness}~\cite{ponce2016chalearn}.
This is advantageous in the sense that personality states can be represented
at any level of the aforementioned big-five personality traits. A Deep Bimodal Regression framework based on both video and audio input was utilized in
\cite{zhang2016deep} to identify personality.
A similar work from G{\"u}{\c{c}}l{\"u}t{\"u}rk \etal\cite{guccluturk2016deep}
introduced a deep audio-visual residual network for
multimodal personality trait recognition.
In addition, a volumetric convolution and Long-Short-Term-Memory (LSTM)
based network was introduced by Subramaniam \etal \cite{subramaniam2016bi}
for learning audio-visual temporal patterns.
A pre-trained CNN was employed by G{\"u}rp{\i}nar \etal \cite{gurpinar2016combining}
to extract facial expressions as well as ambient information
for personality analysis.
For more related work on personality analysis,
please refer to recent surveys~\cite{junior2018first,escalante2018explaining}.

\myPara{Age Estimation.}
Age estimation from face images is gaining its popularity since
the pioneering work of ~\cite{geng2007automatic}.
Conventional regression methods include but are not limited to
kernel method~\cite{guo2009human,guo2011simultaneous},
hierarchical regression~\cite{han2015demographic},
randomized trees~\cite{montillo2009age},
label distribution~\cite{geng2013facial}, and so on.
Recently, end-to-end learning with CNN has also been widely studied for
age estimation.
Ni~\etal\cite{yi2014age} firstly proposed a four layer CNN for age estimation.
Niu~\etal\cite{niu2016ordinal} reformulated age estimation as
an ordinal regression problem by using end-to-end deep learning methods.
In particular, age estimation in their setting was transformed
into a series of binary classification sub-problems.
Ranking CNN was introduced in ~\cite{chen2017using},
where each base CNN was trained with ordinal age labels.
In~\cite{agustsson2017anchored}, anchored Regression Networks were introduced
as a smoothed relaxation of a piece-wise linear regressor for age estimation
through the combination of multiple linear regressors over soft assignments
to anchor points.
Li~\etal\cite{li2018deep} designed a Deep Cross-Population (DCP)
age estimation model with a two-stage training strategy,
in which a novel cost-sensitive multitask loss function was first used
to learn transferable aging features by training on the source population.
Then, a novel order-preserving pair-wise loss function was utilized
to align the aging features of the two populations.
DEX~\cite{rothe2018deep} solved age estimation by way of deep classification
followed by a softmax expected value refinement.
Shen~\etal\cite{Shen_2018_CVPR} extended the idea of a randomized forest
into deep scenarios and show remarkable performances for age estimation.

\myPara{Single Image Super-resolution.}
With the far-reaching application in medical imaging, satellite imaging,
security and surveillance,
single image super-resolution is a classic computer vision problem,
which aims to recover a high resolution (HR) image from a low-resolution (LR) image.
Since the using of fully convolutional networks for super-resolution~\cite{dong2016image},
many advanced deep architectures have been proposed.
For instance, Cascaded Sparse Coding Network (CSCN)~\cite{wang2015deep} combined
the strengths of sparse coding and deep network. An efficient sub-pixel convolution
layer was introduced in~\cite{shi2016real} to better upscale
the final LR feature maps into the HR output. A PCA-inspired collaborative representation cascade was introduced in~\cite{zhang2017collaborative}. A novel residual dense network(RDN) was designed~\cite{zhang2018residual} to fully exploit
the hierarchical features from all the convolutional layers.
Specifically, they proposed residual dense block (RDB) to extract abundant local features via dense connected convolutional layer.
A deeply-recursive convolutional network (DRCN) was proposed in~\cite{kim2016deeply},
which increased the network depth by a recursive layer
without introducing new parameters for additional convolutions.
A very deep fully convolutional encoding-decoding framework was employed
in~\cite{mao2016image} to combine convolution and deconvolution.
Wei \etal\cite{han2018image} reformulated image super-resolution as
a single-state recurrent neural network (RNN) with finite unfoldings
and further designed a dual-state design,
the Dual-State Recurrent Network (DSRN).
Deep Back-Projection Networks (DBPN)~\cite{haris2018deep} exploited iterative
up- and downsampling layers to provide an error feedback message
for projection errors at each stage.  \zl{In ~\cite{zhang2018image}, a residual in residual (RIR) structure was introduce. The concept of non-local learning was adopted in~\cite{zhang2019residual} for image super-resolution.} For more research work, please refer to~\cite{timofte2017ntire}.

% ======================= Proposed Method ============================
\section{Proposed Method}

\subsection{Background}

Before elaborating the proposed regression method,
we first briefly present the notations and the background knowledge.
%to put our method in a proper context.
We assume that we have access to $\nsamples$ training samples,
$\data = \{\samp_1, \dots, \samp_\nsamples \}$.
\zl{The samples are $\dimn$ dimensional: $\data \in \featSpace \subseteq \RR^{\dimn}, \dimn =\dimH \times \dimW \times \dimC$,
where $\dimH,\dimW$ and $\dimC$ denote the height, width and number of channels of
input image respectively.}
Our objective is to predict their regression labels, \ie,
$\Ymat= \{\laby_1,\dots, \laby_\nsamples \}$,
\zl{where  $\Ymat \in \lSpace \subseteq \RR^{\dimny} $}.
We denote a generic data point by $\samp$ and use $\samp_{\diamond}$,
with  $\diamond$ denoting the placeholder for the index wherever necessary.
Similarly, we use $\dimn$ and $\dimny$ to represent the dimensionality
of a generic input data and its label, respectively.
\zl{We achieve our goal by learning a mapping function
$\mapping: \data \rightarrow \lSpace$, where $\mapping \in \mappingfunc$.}

%In a typical regression ensemble,
%dimensionality of input data are considered to be the same, \ie, $\dimn_i=\dimn$.
%In the same way, $\dimny_i=\dimny$.
The learning problem is to use the set $\data$ to learn a mapping function $\mapping$,
parameterized by $\params$, to approximate their label $Y$ as accurate as possible:
\begin{equation} \label{eq:int}
  L(\mapping)=\int (\mapping(\data,\params)-Y)^2p(\data,Y)d(\data,Y),
\end{equation}
In practice, as data distribution $p(X,Y)$ is unknown,
\equref{eq:int} is usually approximated by
\begin{equation} \label{eq:ls}
  L(\mapping)=\frac{1}{\nsamples}
    \sum_{i=1}^{\nsamples}(\mapping(\samp_i,\params)-y_i)^2.
\end{equation}
We omit the input and parameter vectors. Without ambiguity,
instead of $\mapping(\data,\params)$, we write simply $\mapping$.
We use the shorthand expectation operator $E{·}$ to represent
the generalization ability on testing data.
\emph{Bias-variance decomposition}~\cite{brown2005managing} theorem states that
the regression error of a predictor can be decomposed into its
\emph{bias} $\mathbb{B}$ and \emph{variance} $\mathbb{V}$:
\begin{equation}
  E[(\mapping-Y)^2]=\underbrace{{(E[\mapping]-Y)^2}}_{\mathbb{B}(\mapping)^2}
         +\underbrace{E[(\mapping-E[\mapping])^2]}_{\mathbb{V}(\mapping)}.
\end{equation}
It is a property of the generalization error in which bias and variance
have to be balanced against each other for best performance.

A single model, however, turns out to be far from optimal in practice
which has been evidenced by several studies, both
theoretically~\cite{ren2016ensemble,brown2005managing} and
empirically~\cite{fernandez2014we,zhang2017benchmarking}.
Consider the ensemble output $\tilde{\mapping}$ by averaging individual's
response $\mapping_k$, \ie,
\begin{equation} \label{ensemble}
  \tilde{\mapping}=\frac{1}{\nTrees}\sum_{k=1}^{\nTrees}\mapping_k.
\end{equation}
Here we restrict our analysis to the uniform combination case
which is commonly used in practice,
although the decomposition presented below generalize to
non-uniformly weighted ensembles as well.
Posing the ensemble as a single learning unit,
its bias-variance decomposition can be shown by the following equation:
\begin{equation} \label{eqn:bvens}
  E[(\tilde{\mapping}-Y)^2]=
    \underbrace{(E[\tilde{\mapping}]-Y)^2}_{\mathbb{B}(\tilde{\mapping})^2}+
    \underbrace{E[(\tilde{\mapping}-E[\mapping])^2]}_{\mathbb{V}(\tilde{\mapping})}
\end{equation}
Consider ensemble output in \equref{ensemble}, it is straightforward to show:
\begin{equation} \label{bvcen}
\begin{aligned}
  &E[(\tilde{\mapping}-Y)^2]=
    \underbrace{\frac{1}{\nTrees^2}(\sum_{k=1}^{\nTrees}(
    E[\mapping_k]-Y))^2}_{\mathbb{B}({\tilde{\mapping}})^2}\\
  +&\underbrace{\frac{1}{\nTrees^2}\sum_{k=1}^{\nTrees}
    E[(\mapping_k-E[\mapping_k])^2]}_{\mathbb{V}({\tilde{\mapping}})}\\
  +&\underbrace{\frac{1}{\nTrees^2}\sum_{k=1}^{\nTrees}\sum_{j\neq k}
    E[(\mapping_k-E[\mapping_k])
    ({\mapping_j-E[\mapping_j])}]}_{\mathbb{C}({\tilde{\mapping}})},
\end{aligned}
\end{equation}
where $\mathbb{C}$ denotes for \emph{covariance}.

The \emph{bias-variance-covariance} decomposition in \equref{bvcen} illustrates that,
in addition to the internal bias and variance,
the generalization error of an ensemble  depends on the covariance
between the individuals as well.

\begin{figure*}[t]
  \centering
  \begin{overpic}[width=0.95\linewidth]{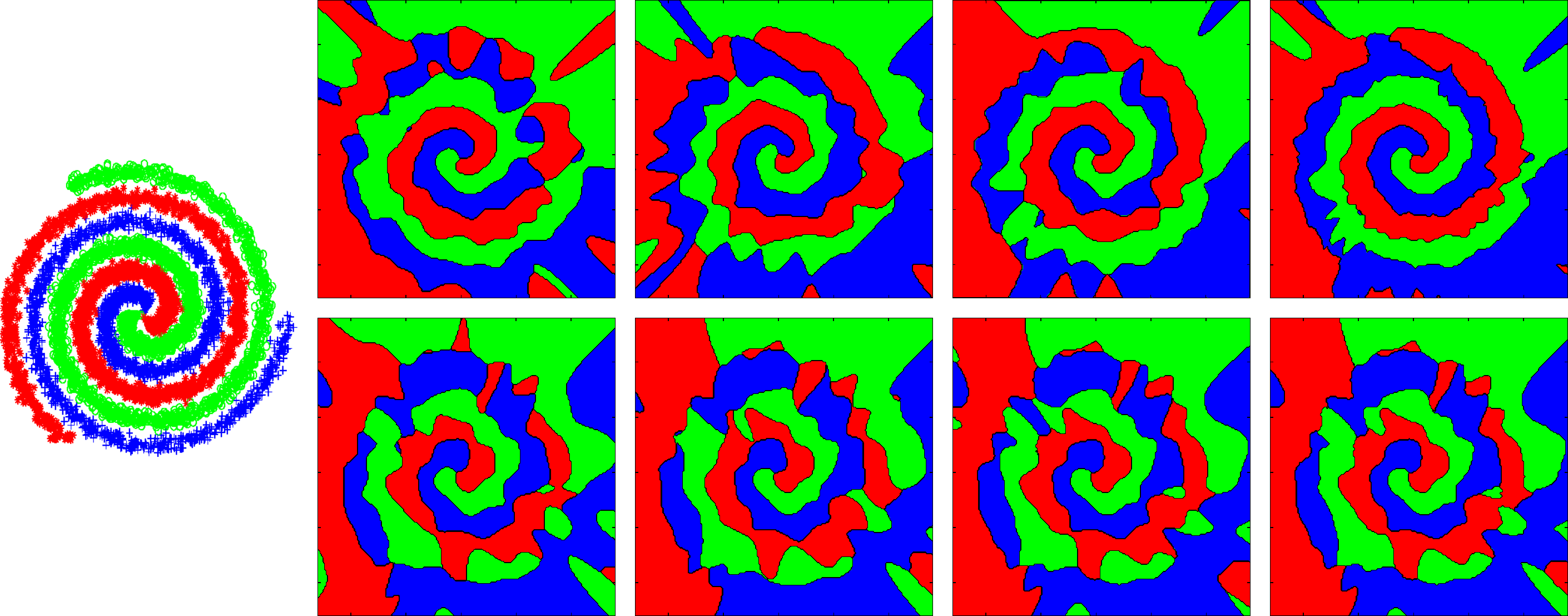}
        \put(2.5,8.8){(a) spirals dataset}
        \put(24.3,-1.8){(b) $1^{st}$ model}
        \put(44.3,-1.8){(c) $2^{nd}$ model}
        \put(64.4,-1.8){(d) $3^{rd}$ model}
        \put(85.5,-1.8){(e) ensemble}
        \put(100.5,26.0){\begin{sideways} (ii) NCL  \end{sideways}}
        \put(100.5,3.2){\begin{sideways} (i) conventional \end{sideways}}
  \end{overpic} \\
  \caption{\zl{Decision surfaces for classification of artificial spirals dataset
     for both (i) conventional ensemble learning and (ii) NCL learning.
     The shading of the background shows the decision surface for that particular class.
     The upper part of the figure corresponds to NCL learning and
     lower part stands for conventional ensemble learning.
     (b)-(d) shows the decision surface of individual model
     and (e) shows the  ensemble decision surface arising from
     averaging over its individual models.
     NCL in (b)-(e) leads to much diversified decision surface
     where errors from individual models may cancel out thus resulting in
     much better generalization ability.
     Best viewed in color.}
  }\label{fig:NCL_DB}
\end{figure*}

\newcommand{\addImg}[1]{\includegraphics[width=0.243\textwidth]{#1}}
\begin{figure}[t]
  \centering
  \subfloat[][Conventional ensemble learning]{\addImg{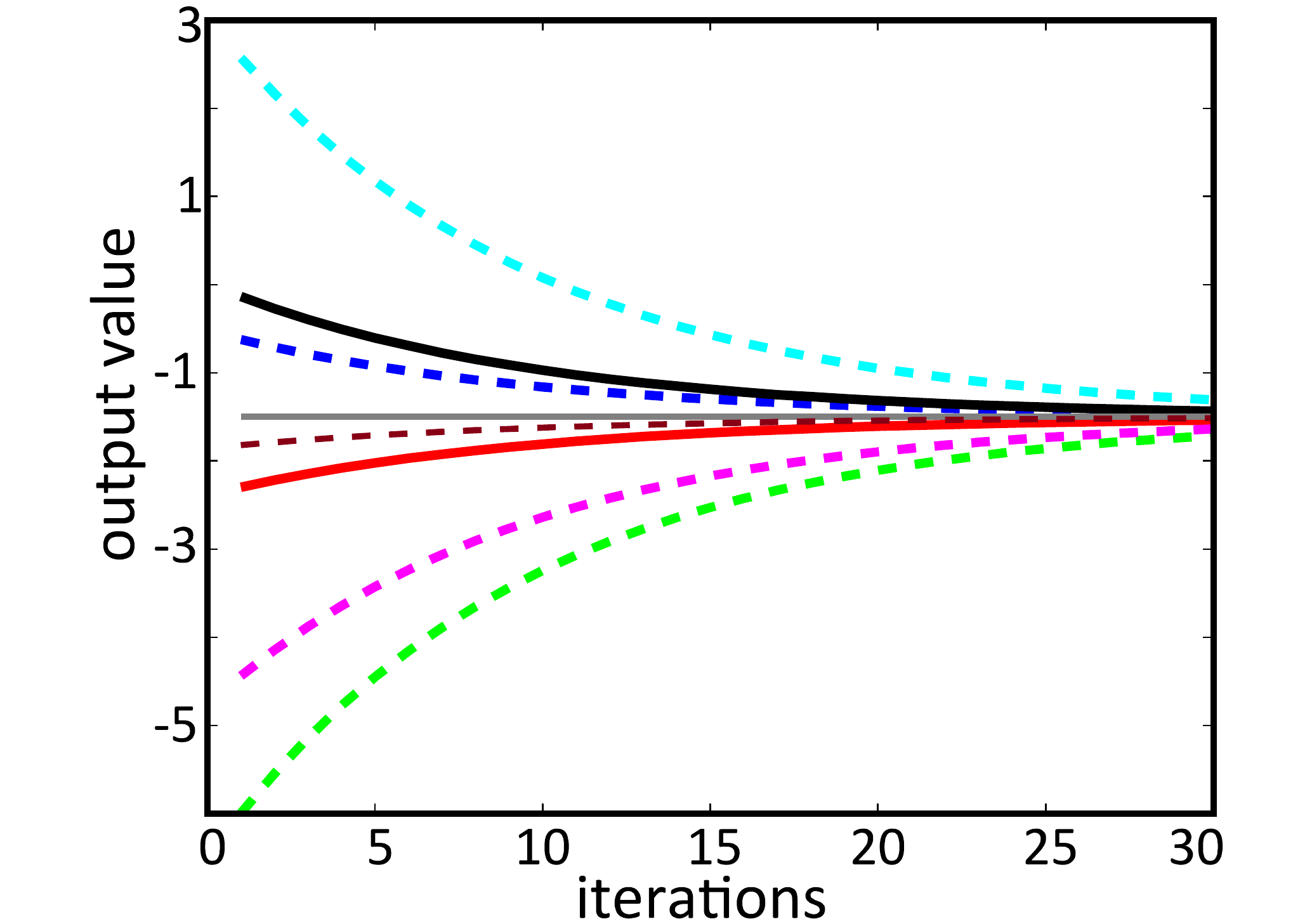}\label{ncl1}}
  \subfloat[][NCL]{\addImg{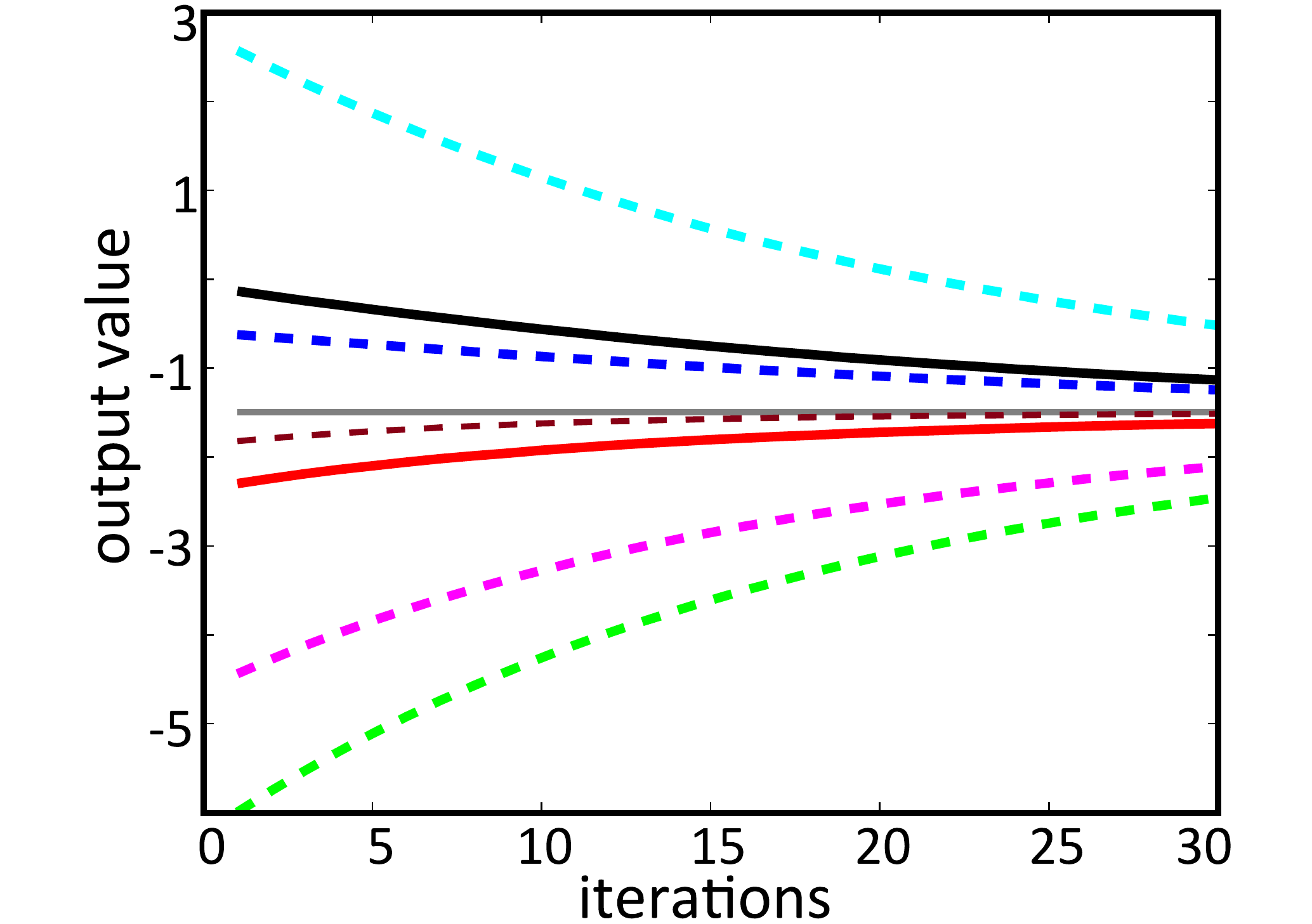}\label{ncl2}}\\
  \caption{Demonstration of the training process of conventional ensemble learning
     and NCL. \zl{The central solid gray line represents the ground truth and all other lines stand for different base models. Although both conventional ensemble learning (left part) and NCL (right part)
may lead to correct estimations by simple model averaging, NCL results in much diversified individual models which make error cancellation being possible on testing data. }
  }\label{steady_state}
\end{figure}

It is natural to show 
\begin{equation} \nonumber
\begin{aligned}
  \frac{1}{\nTrees}\sum_{k=1}^{\nTrees}E[(\mapping_k-Y)^2]= & \mathbb{B}({\mapping})^2  +
     [\nTrees\times \mathbb{V}({\mapping})  \\
   & + \frac{1}{\nTrees}\sum_{k=1}^{\nTrees}(E[\mapping_k]
    - E[\tilde{\mapping}])^2]\\
\end{aligned}
\end{equation}
and
\begin{equation} \nonumber
\begin{aligned}
  \frac{1}{\nTrees}\sum_{k=1}^{\nTrees}E[(\mapping_k-\tilde{\mapping})^2] =
     & -[\mathbb{V}({\mapping})+\mathbb{C}({\mapping})]
     +[\nTrees\times\mathbb{V}({\mapping})\\
  & +\frac{1}{\nTrees}\sum_{k=1}^{\nTrees}(E[\mapping_k]-E[\tilde{\mapping}])^2].\\
\end{aligned}
\end{equation}
Then it is easy to show
\begin{equation} \label{eq:ncl1}
\begin{aligned}
  E[(\tilde{\mapping}-Y)^2] = &
    \frac{1}{\nTrees}\sum_{k=1}^{\nTrees}E[(\mapping_k-Y)^2] \\
    & -\frac{1}{\nTrees}\sum_{k=1}^{\nTrees}E[(\mapping_k-\tilde{\mapping})^2].
\end{aligned}
\end{equation}
\equref{eq:ncl1} explains the effect of error correlations in an ensemble model
by stating that
\emph{the quadratic error of the ensemble estimator is guaranteed to be less than
or equal to the average quadratic error of the component estimators}.
This is also in line with the strength-correlation theory~\cite{breiman2001random},
which advocates learning a set of both accurate and decorrelated models.

\subsection{Deep Negative Correlation Learning}
\subsubsection{\textbf{Our Method}} 
Conventional ensemble learning methods such as bagging~\cite{breiman1996bagging} and Random Forest~\cite{breiman2001random} train multiple models independently. This may not be optimal because, as demonstrated in \equref{eq:ncl1}, the ensemble error consists of both the individual error and the interactions within the ensemble. Based on this, we proposed a
``divide and conquer" deep learning approach by learning a correlation regularized ensemble
on top of deep networks with the following objective:
\begin{equation}\label{NCL}
\begin{aligned}
  L_k &= \frac{1}{2}(\mapping_k-\Ymat)^2+
    \lambda (\mapping_k-\tilde{\mapping})
    (\sum_{j~\neq i}(\mapping_j-\tilde{\mapping})),\\
  &=\frac{1}{2}(\mapping_k-\Ymat)^2-\lambda (\mapping_k-\tilde{\mapping})^2,
\end{aligned}
\end{equation}

More specifically, we consider our mapping function as an ensemble of predictors
as defined in \equref{ensemble} where each base predictor is posed as:
\begin{equation}
\begin{aligned}
  \mapping_k(\samp_i)
  &=\mapping^{\nStage}_k(\mapping^{\nStage-1}_k\cdots(\mapping^1_k(\samp_i)))),\\
  &k= 1,2\cdots\nTrees, i=1,2\cdots\nsamples
\end{aligned}\label{crowd_counting}
\end{equation}
\zl{where $k$,   $i$, and} $\nStage$ \zl{stand for the index for individual models, the index for data samples and the depth of the network, respectively.}
More specifically, each predictor in the ensemble consists of
cascades of feature extractors $\mapping^q_k$, $q=1,2\cdots\nStage-1$
and regressor $\mapping^{\nStage}_k$.
Motivated by the recent success of CNNs on visual recognition tasks,
each feature extractor $\mapping^q_k$ is embodied by a typical layer of a CNN. Below we present the details for each task.

% By ``Fully Convolutional" we mean $\mapping$ commutes with translation. More formally, considering a translation operator  for a one-dimensional  signal $(T_\kappa\samp)[i] = \samp[i-\kappa]$, a function  $\mapping$ that maps
% signals to signals is fully-convolutional with integer stride $s$ if
% $\mapping(T_{\kappa s}\samp)=T_\kappa\mapping(\samp)$ for any translation $\kappa$. The definition generalize to image signals in a straightforward way.  Fully-convolutional realization of $\mapping$ is advantageous in the sense that  one can provide input in an arbitrary size to the network and it will compute mapping results, which is the estimation of crowd density map.

\begin{figure*}
    \centering
    \includegraphics[width=0.95\textwidth]{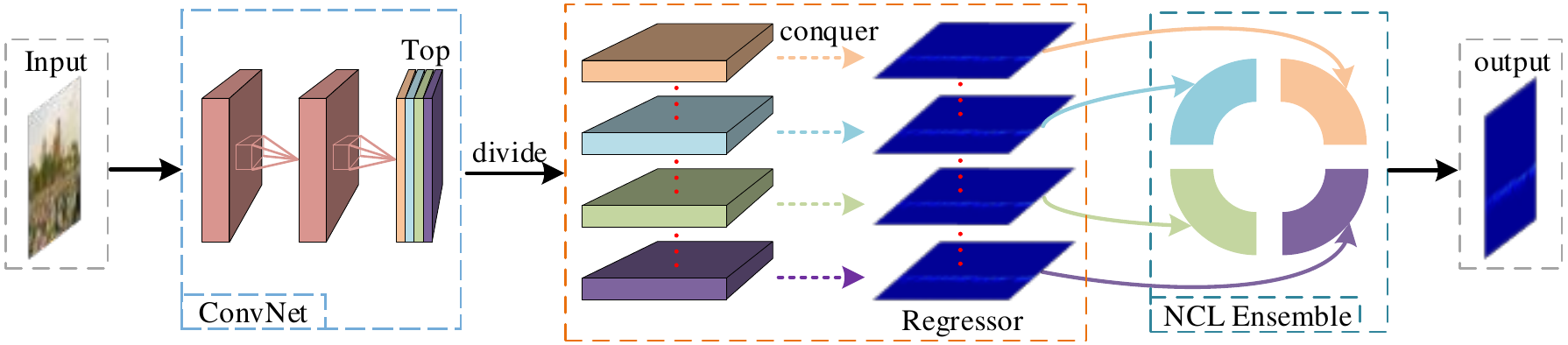}
    \caption{\zl{Details of the proposed DNCL. Regression is formulated as ensemble learning with the same amount of parameter as a single CNN. DNCL  processes the input by a stack of typical convolutional and pooling layers. Finally, a ``divide and conquer" strategy is adapted to learn a pool of regressors to regress the output on top of each convolutional feature map at top layers. Each regressor is jointly optimized with the CNN by an amended cost function which penalizes correlations with others to make better trade-offs among the bias-variance-covariance in the ensemble. }}
    \label{fig:system}
\end{figure*}

\subsubsection{\textbf{Network Structure}} 
The proposed method can be efficiently encapsulated into existing deep CNN thanks to its rich feature hierarchy.  In our implementation, as illustrated in \figref{fig:system},
lower levels of feature extractors are shared by each predictor for efficiency,
\ie, $\mapping^q_k=\mapping^q$, $q=1,2\cdots,\nStage-1$, $k=1,2\cdots, \nTrees$.
Furthermore, building on the lessons learnt from subspace idea
in ensemble learning~\cite{breiman2001random},
highest level of feature extractor $\mapping^{\nStage-1}_k$ outputs a different
feature subset for different regressor $\mapping^{\nStage}_k$ to insert more diversities. In this study, this is implemented via the well-established ``group convolution" strategy~\cite{krizhevsky2012imagenet}. Each regressor is optimized by  an amended cost function as defined in Eqn.~\ref{NCL}.
Generally speaking, network specification of $\mapping^q_k$ is problem dependent
and we show that,
the proposed method is end-to-end-trainable and independent of
the backbone network architectures.

\myPara{Crowd counting.} 
We employ a deep pretrained VGG16 network for this task and
make several modifications.
Firstly, the stride of the fourth max-pool layer is set to 1.
Secondly, the fifth  pooling  layer was further removed.
This provides us with a much larger feature map with richer information.
To handle the receptive-field mismatch caused by the removal of stride
in the fourth max-pool layer,
we then double the receptive field of convolutional layers after
the fourth max-pool layer by using the technique of \textit{holes} 
introduced in~\cite{yu2015multi}.
We also include another variant of the proposed method called
``NCL" which is a shallow network optimized with the same loss function.
The details of this network will be elaborated in \secref{sec:network}.

\myPara{Personality analysis.} We utilize a truncated 20 layer version of the \emph{SphereFace}
model~\cite{liu2017sphereface} for personality analysis.
We first detect and align faces for each input image
with well-established \emph{MTCNN}~\cite{zhang2016joint}. As we are dealing with videos,
in order to speed up training and reduce the risk of over-fitting,
we take a similar approach as done in~\cite{wang2016temporal}
to first sparsely sample 10 frames from each video in a randomized manner.
Average pooling is further used to aggregate multiple results
for the same video.

\myPara{Age estimation.} For age estimation, we use the network backbone of deep forest~\cite{Shen_2018_CVPR}.
It reformulates the split nodes of a decision forest as a fully connected layer
of a CNN and learns both split nodes and leaf nodes in an iterative manner.
More specifically, by fixing the leaf nodes, the split nodes as well as the CNN parameters are optimized by back-propagation.
Then, by fixing the split nodes, the leaf nodes are optimized by
iterating a step-size free and fast converging update rule
derived from Variational Bounding.
Instead of using this iterative strategy,
we use the proposed NCL loss in each node to make them both accurate
and diversified.

\myPara{Image super-resolution.} For image super-resolution, we choose the \sArt DRRN~\cite{tai2017image}
as our network backbone and change the L2 loss into the proposed NCL loss.
More specifically, an enhanced residual unit structure is recursively learned in
a recursive block, and  several recursive blocks are stacked to
learn the residual image between the HR and LR images.
The residual image is then added to the input LR image
from a global identity branch to estimate the HR image.

\equref{NCL} can be regarded as a smoothed version of \equref{eq:ncl1}
to improve the generalization ability of the ensemble models.
Please note that the optimal value of $\lambda$ may not necessarily
be 0.5 because of the discrepancy between the training and testing
data~\cite{brown2005managing}.
By setting $\lambda=0$, we actually achieve conventional ensemble learning
(non-boosting type) where each model is optimized independently.
It is straightforward to show that the first part in \equref{NCL} corresponds
to bias plus an extra term
$[\nTrees\times\mathbb{V}({\mapping})+\frac{1}{\nTrees}
\sum_{k=1}^{\nTrees}(E[\mapping_k]-E[\tilde{\mapping}])^2]$,
while the second part stands for the variance, covariance and the same term
$[\nTrees\times\mathbb{V}({\mapping})+\frac{1}{\nTrees}
\sum_{k=1}^{\nTrees}(E[\mapping_k]-E[\tilde{\mapping}])^2]$.
Since the extra term appears on both sides,
it cancels out when we combine them by subtracting, as done in \equref{NCL}.
Thus by introducing the second part in \equref{NCL},
we aim at achieving better ``diversity" with negative correlated base models
to balance the components of bias variance and the ensemble covariance
to reduce the overall mean square error (MSE).

To demonstrate this, consider the scenario in \figref{steady_state}.
We are using a regression ensemble consisting of $6$ regressors
where the ground truth is $-1.5$.
Each curve in \figref{steady_state} illustrates the evolution of one regressor
when trained with gradient descent, \ie,
$f_{i,n}=f_{i,n-1}-\gamma \frac{d_E}{d_{f_{i,n-1}}}$,
where $\gamma$ and $E$ stands for the learning rate and mean-square loss function,
respectively.
$i\in \{1,2,\cdots, 6\}$ is the index of individual models in the ensemble and
$n\in \{1,2,\cdots, 30\}$ stands for the index of iterations.
Although both conventional ensemble learning (\figref{ncl1}) and
NCL (\figref{ncl2}) may lead to correct estimations by simple model averaging,
NCL results in much diversified individual models
which make error cancellation being possible on testing data.

For generalization, consider the artificial spirals dataset in \figref{fig:NCL_DB}(a),
where an ensemble of three single hidden layer feed-forward network (SLFN)
is trained on.
Then the ensemble is evaluated on data samples densely sampled on $x-y$  plane.
The first row in \figref{fig:NCL_DB} shows that NCL ensemble leads to
more diversified SLFN,
compared with conventional ensemble learning as illustrated in
the second row of \figref{fig:NCL_DB},
thus making the resulting ensemble generalize well on testing data.
\zl{Creating diverse sets of models has been extensively studied,
both theoretically
\cite{brown2005managing,brown2005diversity,ren2016ensemble,dietterich2000ensemble,minku2009impact,lee2016stochastic,alhamdoosh2014fast,zhou2002ensembling}
and empirically~\cite{fernandez2014we,hansen1990neural}.}
More specifically, Breiman~\cite{breiman2001random} derived a VC-type bound
for generalization ability of ensemble models
which advocated both accurate and decorrelated individual models. In addition, our methods also differ from the classical work of~\cite{liu2000evolutionary} which trains multiple shallow networks.

\subsubsection{\textbf{Connection with the Rademacher Complexity}} We now show the bound for the Rademacher complexity~\cite{bartlett2002rademacher}
of the proposed deep negatively correlation learning.
Firstly we will make no difference between convolution and
fully-connected (FC) layers because FC layers can be easily transformed
into convolution layers with property kernel size and padding values.
% \begin{figure}[!thb]
%     \centering
%     \includegraphics[width=0.9\columnwidth]{image/conv2fc.jpg}
%     \label{fig:conv2fc}
%     \caption{An illustration of the equivelence between convolution and fully-connected batch matrix multiplication.}
% \end{figure}

\mypartitletwo{Definition 1.} (Radamacher Complexity).
\emph{For a dataset $\data = \{\samp_1,\dots, \samp_\nsamples \}$
generated by a distribution $\distribution$ on set $\featSpace$ and
a real-valued function class $\mappingfunc$ in  $\featSpace$,
the empirical Rademacher complexity of $\mapping$ is the random variable:}
\begin{equation}
  \hat{R}_{\nsamples}(\mappingfunc)=E_{\sigma} \begin{bmatrix}
  \underset{\mapping \in \mappingfunc}{\sup}
  |\frac{2}{\nsamples}\sum_{i=1}^{\nsamples}\sigma_i\mapping(\samp_i)|
\end{bmatrix}\label{empirical_rade},
\end{equation}
\emph{where $\sigma_1,\cdots,\sigma_{\nsamples}$ are usually referred as
Rademacher Variable and are independent random variables uniformly chosen
from $\{-1,+1\}$.
The Rademacher complexity of $\mappingfunc$ is
$R_{\nsamples}(\mappingfunc)=E_{\data}[\hat{R}_{\nsamples}(\mappingfunc)]$.}

The empirical Rademacher complexity is widely regarded as a proximity
of the generalization ability based on the following theorem:

\mypartitletwo{Theorem 1.}
(Koltchinskii and Panchenko, 2000~\cite{koltchinskii2002empirical}).
\emph{Fix $\delta  \in (0,1)$ and let $\mappingfunc$ be a class of
functions mapping from $\data$ to [0,1].
Let $\samp_i \in \data$ be drawn independently according to
a probability distribution $\distribution$.
Then with probability at least $1-\delta$ over random draws
of samples of size $\nsamples$, every $\mapping \in \mappingfunc$ satisfies:}
\begin{equation}
\begin{aligned}
  E[\mapping(\data)]
  \leq \hat{E}[\mapping(\data)]
  +\hat{R}_{\nsamples}(\mappingfunc)
  +3\sqrt{\frac{\ln(\frac{2}{\delta})}{2\nsamples}}
\end{aligned}
\end{equation}
\emph{where $\hat{E}[\mapping(\data)]
=\frac{1}{\nsamples}\sum_{i=1}^{\nsamples}\mapping(\samp_i)$}.

\vspace{5pt}
In addition, we have

\mypartitletwo{Lemma 1.}
\emph{For $\mappingfunc$ and $\phi: \mathbb{R} \rightarrow \mathbb{R}$, let $\mappingfunc':=\{\phi \circ \mapping: \mapping \in \mappingfunc\}$. If $\phi$ is $\loss$-Lipschitz continuous, i.e.  $|\phi(x)-\phi(x')| \leq 
\zl{\loss} |x-x'|$, then for any $\nsamples$:}
\begin{equation}
    \hat{R}_{\nsamples}(\mappingfunc') \leq \loss \hat{R}_{\nsamples}(\mappingfunc)
    \label{equ:lemma1}
\end{equation}

\mypartitletwo{Proof.}
We provide the proof for $\nsamples=1$, the general case works iteratively.
\begin{equation}
\begin{aligned}
  \hat{R}_{\nsamples}(\mappingfunc')
  &=E_{\sigma_1}\begin{bmatrix}
    \underset{\mapping \in \mappingfunc}{\sup}(2\sigma_1\phi(\mapping(\samp_1)))
    \end{bmatrix}\\
  &=\underset{\mapping \in \mappingfunc}{\sup}(\phi(\mapping(\samp_1)))
    -\underset{\mapping \in \mappingfunc}{\sup}(\phi(\mapping(\samp_1)))\\
  &=\underset{\mapping^1 \in \mappingfunc, \mapping^2 \in
    \mappingfunc}{\sup}(\phi(\mapping^1(\samp_1))-\phi(\mapping^2(\samp_1)))\\
  &\leq \underset{\mapping^1 \in \mappingfunc, \mapping^2 \in
    \mappingfunc}{\sup}(\loss|\mapping^1(\samp_1)-\mapping^2(\samp_1)|)\\
  &=\loss(\underset{\mapping^1 \in \mappingfunc}{\sup}(\mapping^1(\samp_1))+
    \underset{\mapping^2 \in \mappingfunc}{\sup}(-\mapping^2(\samp_1)))\\
  &=\loss E_{\sigma_1}\begin{bmatrix}
    \underset{\mapping \in \mappingfunc}{\sup}(2\sigma_1\mapping(\samp_1))
    \end{bmatrix}\\
  &=\loss \hat{R}_{\nsamples}(\mappingfunc)
\end{aligned}
\end{equation}

Based on Lemma 1, we have the following conclusion:

\mypartitletwo{Lemma 2.}
\emph{Let  $p\geq1$, $Z=\data\times Y$, $L_p(z)=L_p(\samp,\laby)
=\{\samp \rightarrow |\mapping(\samp)-\laby|^p: \mapping \in \mappingfunc\}$.
Assume that $\begin{bmatrix} \underset{\mapping \in \mappingfunc,\samp
\in \data}{\sup}|\mapping(\samp)-\laby| \end{bmatrix}\leq M$.
Then for any sample $\data$ of size $\nsamples$:}
\begin{equation}\label{theorem1}
  \hat{R}_{\nsamples}(L_p) \leq pM^{p-1}\hat{R}_{\nsamples}(\mapping).
\end{equation}

\mypartitletwo{Proof} let $L'=\{\samp\rightarrow \mapping(\samp)-\laby: \mapping\in\mappingfunc\}$. Then we have $L_p=\{\phi\circ l: l\in L'\}$
with $\phi: \samp\rightarrow|\samp|^p$.
From Lemma 1, we have $\phi$ is $pM^{p-1}$ Lipschitz over $[-M,M]$, then we have:
\begin{equation}
    \hat{R}_{\nsamples}(L_p)\leq pM^{p-1}\hat{R}_{\nsamples}(L').
    \label{equ:loss_complexity}
\end{equation}
Moreover, with
\begin{equation}
\begin{aligned}
  \hat{R}_{\nsamples}(L')&=E_{\sigma} \begin{bmatrix}
    \underset{\mapping \in \mappingfunc}{\sup}(\frac{2}{\nsamples}
    \sum_{i=1}^{\nsamples}\sigma_i\mapping(\samp_i)+\sigma_i y_i)
    \end{bmatrix}\\
  &=E_{\sigma} \begin{bmatrix}
   \underset{\mapping \in \mappingfunc}{\sup}(\frac{2}{\nsamples}
   \sum_{i=1}^{\nsamples}\sigma_i\mapping(\samp_i))
   \end{bmatrix}+E_{\sigma} \begin{bmatrix}
   (\frac{2}{\nsamples}\sum_{i=1}^{\nsamples}\sigma_i y_i)
   \end{bmatrix}\\
  &=\hat{R}_{\nsamples}(\mapping),
\end{aligned}
\end{equation}
we complete the proof.

Furthermore, by combining $\mapping(x)=L_2$ as defined in Lemma 2
with Theorem 1, we have

\mypartitletwo{Lemma 3.}
\emph{Let $\data = \{\samp_1,\dots, \samp_\nsamples \}$
be a dataset generated by a distribution $\distribution$ on set $\featSpace$
and  $\Ymat= \{\laby_1,\dots, \laby_\nsamples \}$ be their corresponding labels.
For a function class $\mappingfunc \subseteq \{\mapping: \data \rightarrow \Ymat\}$
which maps the data $\data$ to [0,1],
define the commonly used mean square as
$\ell_{\distribution}(\mapping)=E_{\distribution}[(\mapping(x)-y)]^2$
and the empirical mean square error as as
$\hat{\ell}_{\data}(\mapping)=E_{\data}[(\mapping(x)-y)^2]$.
Assume that $\begin{bmatrix}
\underset{\mapping \in \mappingfunc,\samp \in \data}{\sup}|\mapping(x)-\laby|
\end{bmatrix}\leq M$.
Then for a fixed $\delta  \in (0,1)$, with probability at least
$1-\delta$ over random draws of samples of size $\nsamples$,
every $\mapping \in \mappingfunc$ satisfies}

\begin{equation}
    \begin{aligned}
    \ell_{\distribution}(\mapping)\leq \hat{\ell}_{\data}(\mapping)+2M\hat{R}_{\nsamples}(\mappingfunc)+\sqrt{\frac{ln{\frac{2}{\delta}}}{2\nsamples}}.
    \end{aligned}
\end{equation}

% Let $Z=\data\times Y$, $L_2(z)=L_2(\samp,\laby)=\{\samp \rightarrow |\mapping(\samp)-\laby|^2: \mapping \in \mappingfunc\}$. Assume that $\begin{bmatrix}
% \underset{\mapping \in \mappingfunc,\samp \in \data}{\sup}|\mapping(x)-\laby|
% \end{bmatrix}\leq M$. Then for any sample $\data$ of size $\nsamples$:

We now show that the empirical Rademacher complexity of the proposed method
on the training set is $\frac{1}{\nTrees}$ to the standard network:

\mypartitletwo{Proposition 1.}
Denote by  $\tilde{\mapping} \in \tilde{\mappingfunc}$
the group convolution based method and by ${\mapping} \in {\mappingfunc}$,
the conventional method. Then
\begin{equation}\label{equ:Proposition1}
  \hat{R}_{\nsamples}(\tilde{\mappingfunc})
  =\frac{1}{\nTrees}\hat{R}_{\nsamples}(\mappingfunc).
\end{equation}

\mypartitletwo{Proof.}
\begin{equation}
\begin{aligned}
  \hat{R}_{\nsamples}(\tilde{\mappingfunc})
  &=E_{\sigma} \begin{bmatrix}
    \underset{\tilde{\mapping} \in \tilde{\mappingfunc}}{\sup}|\frac{2}{\nsamples}
    \sum_{i=1}^{\nsamples}\sigma_i\tilde{\mapping}(\samp_i)|\end{bmatrix}\\
  &=E_{\sigma} \begin{bmatrix}
    \underset{\tilde{\mapping} \in
    \tilde{\mappingfunc}}{\sup}|\frac{2}{\nsamples*\nTrees}
    \sum_{i=1}^{\nsamples}\sigma_i{\sum_{k=1}^{\nTrees}
    \tilde{\mapping}_k(\samp_i)}|\end{bmatrix}\\
  &=E_{\sigma} \begin{bmatrix}
    \underset{\tilde{\mapping} \in
    \tilde{\mappingfunc}}{\sup}|\frac{2}{\nsamples*\nTrees}
    \sum_{i=1}^{\nsamples}\sigma_i\sum_{k=1}^{\nTrees}
    \tilde{\mapping}^{\nStage-1}_k(\samp_i)\circledast W^{\nStage}_k| \end{bmatrix}\\
  &=E_{\sigma} \begin{bmatrix}
    \underset{\mapping \in
    \mappingfunc}{\sup}|\frac{2}{\nsamples*\nTrees}
    \sum_{i=1}^{\nsamples}\sigma_i{\mapping(\samp_i) \circledast W^\nStage}|\end{bmatrix}\\
  &=\frac{1}{\nTrees}\hat{R}_{\nsamples}(\mappingfunc).
\end{aligned}\label{rademacher_group}
\end{equation}
\emph{The operation} $\sum_{k=1}^{\nTrees}\mapping^{\nStage-1}_k(\samp_i)\circledast W_k$
\emph{in \equref{rademacher_group} stands for the convolution operator.}
$\mapping^{\nStage-1}_k(\samp_i)$ \emph{stand for the  $k^{th}$ feature subset
of the feature maps} $\mapping^{\nStage-1}(\samp_i)$.
\emph{More specifically,
we divide the feature map of} $\mapping^{\nStage-1}(\samp_i) \subseteq
\RR^{\dimH^{\nStage-1}_i \times \dimW^{\nStage-1}_i \times \dimC^{\nStage-1}_i}$
\emph{along the $3^{rd}$ axis into $\nTrees$ subset.
The same procedure is applied on the kernel filter} $W^{\nStage}$.

\mypartitletwo{Remark 1.}
\emph{The empirical Rademacher complexity measures the ability of functions
from a function class (when applied to a fixed set $\data$) to fit random noise.
It is a more modern notion of complexity that is distribution dependent and
defined for any class real-valued functions. On the one hand, by setting $\nTrees>1$,
our method works in a ``divide and conquer" manner and the whole
Rademacher complexity is reduced by a factor of $\nTrees$,
which, intuitively speaking,
making the function $\tilde{\mapping} \in \tilde{\mappingfunc}$ easier to learn.
On the other hand, $\nTrees$ may also affect the term of
$\ell_{\data}(\mapping)$.
For instance, setting an extremely larger value of $\nTrees$
may also lead to a larger value of $\ell_{\data}(\mapping)$
because much less input feature is provided for each base predictor.}

\section{Experiment}

In this section, we investigate the feasibility of the proposed method
on four regression tasks: crowd counting, personality analysis,
age estimation and single image super-resolution.
The proposed method is implemented in Caffe~\cite{jia2014caffe}.
In order to further understand the merits of the proposed methods,
we also include some variants of the proposed method.
More specifically, for each task, we replace the proposed loss function with L2,
SmoothL1 and Tukey loss, and
they are referred as ``L2", ``SmoothL1" and ``Tukey", respectively.
For the SmoothL1 loss, instead of using a fixed value of $1$ for the threshold
in \equref{equ:smoothL1},
we treat it as another hyper-parameter and optimized it on the training data.
We do not compare the proposed method explicitly with naive implementations
of multiple CNN ensemble as their computational time is much larger
and thus are less interested to us.
We highlight the best results in each case in \textbf{\textcolor{red}{Red}}.
The second and third best methods are highlighted in \textcolor{green}{Green}
and \textcolor{blue}{Blue}, respectively.
As different evaluation protocols may be utilized in different applications,
we put a $\uparrow$ after each metric to indicate the cases wherever
a larger value is better.
Similarly, $\downarrow$
is used in cases wherever smaller value indicates better performance.

\newcommand{\First}[1]{\textcolor{red}{\textbf{#1}}}
\newcommand{\Second}[1]{\textcolor{green}{#1}}
\newcommand{\Third}[1]{\textcolor{blue}{#1}}

\subsection{Crowd Counting}
For crowd counting, we evaluate the proposed methods on three benchmark datasets:
UCF\_CC\_50 dataset~\cite{idrees2013multi},
Shanghaitech dataset~\cite{zhang2016single} and
WorldExpo'10 dataset~\cite{zhang2015cross}.
The proposed networks are trained using Stochastic Gradient Descent
with a mini-batch size of 1 at a fixed constant momentum value of 0.9.
Weight decay with a fixed value of 0.0005 is used as a regularizer.
We use a fixed learning rate of $10^{-7}$ in the last convolution layer
of our crowd model to enlarge the gradient signal for effective
parameter updating and use a relatively smaller learning rate
of $10^{-9}$ in other layers. We set the ensemble size to be 64. More specifically, we use a convolution layer with the kernel of
$64 \times 8\times 1\times 1$ as regressor $\mapping^\nStage_k$ on
each output feature map to get the final crowd density map.
Specifically, each regressor $\mapping^\nStage_k$ is sparsely connected
to a small portion of feature maps from the last convolutional layer
($conv5\_3$) of VGG16 network,
implemented via the well-established ``group convolution"
strategy~\cite{krizhevsky2012imagenet,wang2016stct}.
We also include another variant of the proposed method called ``NCL",
which is a shallow network optimized with the same loss function.
The details of this network will be elaborated in \secref{sec:network}.

% We first employ a deep pretrained VGG16 network for this task and
% make several modifications.
% Firstly, the stride of the fourth max-pool layer is set to 1.
% Secondly, the fifth  pooling  layer was further removed.
% This provides us a much larger feature map with richer information.
% To handle the receptive-field mismatch caused by the removal of stride
% in the fourth max-pool layer,
% we then double the receptive field of convolutional layers after
% the fourth max-pool layer by using the technique of holes introduced
% in~\cite{yu2015multi}.
% Finally, we use a convolution layer with the kernel of
% $64 \times 8\times 1\times 1$ as regressor $\mapping^\nStage_k$ on
% each output feature map to get the final crowd density map.
% Specifically, each regressor $\mapping^\nStage_k$ is sparsely connected
% to a small portion of feature maps from the last convolutional layer
% ($conv5\_3$) of VGG16 network,
% implemented via the well-established ``group convolution"
% strategy~\cite{krizhevsky2012imagenet,wang2016stct}.
% We also include another variant of the proposed method called
% ``NCL" which is a shallow network optimized with the same loss function.
% The details of this network will be elaborated in \secref{sec:network}.

The widely used \emph{mean absolute error}~(MAE) and the
\emph{root mean squared error} (RMSE) are adopted to evaluate the performance
of different methods.
The MAE and RMSE are defined as follows:
\begin{equation}
  \rm{ MAE} = \frac{1}{N}\cdot \sum_{i=1}^{N}|(y_i- \tilde{y}_i)|,
\end{equation}
\begin{equation}
  \rm{ RMSE} = \sqrt{\frac{1}{N}\cdot \sum_{i=1}^{N}(y_i- \tilde{y}_i)^2}
\end{equation}
Here $N$ represents the total number of images in the testing datasets,
$y_i$ and $\tilde{y}_i$ are the ground truth and the estimated value
respectively for the $i^{th}$ image.

\newcommand{\RodriguezCC}{Rodriguez \etal \cite{rodriguez2011density}}
\newcommand{\LempitskyCC}{Lempitsky \etal \cite{lempitsky2010learning}}
\newcommand{\IsreesCC}{Isrees \etal \cite{idrees2013multi}}
\newcommand{\ZhangLCC}{Zhang \etal \cite{zhang2015cross}}
\newcommand{\CrowdNet}{CrowdNet \cite{boominathan2016crowdnet}}
\newcommand{\ZhangZCC}{Zhang \etal \cite{zhang2016single}}
\newcommand{\ZengCC}{Zeng \etal \cite{zeng2017multi}}
\newcommand{\MarkCC}{Mark \etal \cite{MarsdenMLO16a}}
\newcommand{\DanielCC}{Daniel \etal \cite{onoro2016towards}}
\newcommand{\SamCC}{Sam \etal \cite{sam2017switching}}
\newcommand{\EladCC}{Elad \etal\cite{walach2016learning}}
\newcommand{\LiuCC}{Liu~\etal\cite{liu2018decidenet}}

\begin{table}[t]
  \caption{Comparing results of different methods on the UCF\_CC\_50 dataset.}
  \centering
  \renewcommand{\tabcolsep}{9pt}
  \begin{tabular}{c|c|c|c} \hline
    Method & Deep Features & MAE~$\downarrow$ & RMSE~$\downarrow$\\\hline
    \RodriguezCC & \xmark & 655.7 & 697.8\\
    \LempitskyCC & \xmark & 493.4 & 487.1\\
    \IsreesCC    & \xmark & 419.5 & 541.6\\
    \ZhangLCC    & \cmark & 467.0 & 498.5\\
    \CrowdNet    & \cmark & 452.5 & -\\
    \ZhangZCC    & \cmark & 377.6 & 509.1\\
    \ZengCC      & \cmark & 363.7 & 468.4\\
    \MarkCC      & \cmark & 338.6 & \Second{424.5}\\
    \DanielCC    & \cmark & \Third{333.7} & \Third{425.2}\\
    \SamCC       & \cmark & \Second{318.1} & 439.2\\
    \EladCC      & \cmark &364.2&-\\
    \hline
    L2           & \cmark &394.3&556.9\\
    SmoothL1     & \cmark &384.1&556.7\\
    Tukey        & \cmark &380.7&552.0\\
    NCL          & \cmark &354.1&443.7\\
   \mymodel      & \cmark &\First{288.4}&\First{404.7}\\
   \hline
  \end{tabular}\label{tab_ucf}
\end{table}

\myPara{UCF\_CC\_50 dataset} The challenging UCF\_CC\_50 dataset \cite{idrees2013multi} contains 50 images
that are randomly collected from the Internet.
The number of head ranges from $94$ to $4543$ with an average of
$1280$ individuals per image.
The total number of annotated persons within $50$ images is $63974$.
Challenging issues such as large variations in head number among different
images from a small amount of training images come in the way of
accurately counting for UCF\_CC\_50 dataset.
We follow the standard evaluation protocol by splitting the dataset
randomly into five parts in which each part contains ten images.
Five-fold cross-validation is employed to evaluate the performance.
Since the perspective maps are not provided,
we generate the ground truth density map by using the method of
Zhang \etal\cite{zhang2016single}.

We compare our method on this dataset with the \sArt methods.
In \cite{rodriguez2011density,lempitsky2010learning,idrees2013multi},
handcraft features are used to regress the density map from the input image.
Several CNN-based methods in \cite{zhang2015cross,boominathan2016crowdnet,
zhang2016single,zeng2017multi,MarsdenMLO16a,onoro2016towards,sam2017switching}
were also considered here due to their superior performance on this dataset.
\tabref{tab_ucf} summarizes the detailed results.
Firstly, it is obvious that most deep learning methods outperform
hand-crafted features significantly.
In~\cite{boominathan2016crowdnet}, Boominathan \etal proposed to
employ a shallow network to assist the training process of deep VGG network.
With the proposed deep negative learning strategy,
it is also interesting to see that
1) both our deep (``DNCL") and shallow (``NCL") networks work well;
2) deep networks (``DNCL") are better than shallower networks (``NCL"),
as expected.
However, shallower network still leads to competitive results and may be
advantageous in resource-constrained scenarios as it is computationally cheaper;
(3) it is straightforward to see that the deeper version of
the proposed method outperforms all others on this dataset;
(4) the proposed method performs favorably against a naive application
of multiple CNN ensemble of~\cite{walach2016learning}.

\newcommand{\MAE}{MAE $\downarrow$}
\newcommand{\RMSE}{RMSE~$\downarrow$}

\begin{table}[t]
  \centering
  \renewcommand{\tabcolsep}{3mm}
  \caption{Comparison of crowd counting methods
    on the Shanghaitech dataset.}
  \begin{tabular}{c|c|c|c|c} \hline
    \RowsT{Method}& \ColsT{|}{Part\_A} & \ColsT{}{Part\_B} \\ \cline{2-5}
              & \MAE  & \RMSE & \MAE &    \RMSE \\\hline
    LBR+RR    & 303.2 & 371.0 & 59.1 & 81.7 \\
    \ZhangLCC & 181.8 & 277.7 & 32.0 & 49.8 \\
    \ZhangZCC & 110.2 & 173.2 & 26.4 & 41.3 \\
    \SamCC    &\Second{90.4}&\Second{135.0}&\Third{21.6}&\Third{33.4}\\
    \LiuCC    &   -   &  -    &\Second{20.8}&\Second{29.4}\\\hline
    L2        & 105.4 & 152.3 & 40.4 & 58.6\\
    SmoothL1  &\Third{94.9}&\Third{150.1}&    40.3    &    58.3 \\
    Tukey     & 104.5 & 151.2 & 40.3 & 58.4 \\
    NCL       & 101.7 & 152.8 & 25.7 & 38.6 \\
    \mymodel  &\First{73.5}&\First{112.3}&\First{18.7}& \First{26.0} \\ \hline
  \end{tabular} \label{tab_tech}
\end{table}

\begin{table}[b]
  \centering
  \renewcommand{\tabcolsep}{7pt}
  \caption{Comparison of mean absolute error (\MAE)
    of different crowd counting methods on the WorldExpo'10 dataset.}
  \begin{tabular}{c|c|c|c|c|c|c} \hline
    \RowsT{Method} &  \multicolumn{5}{c|}{Scenes}
      & \RowsT{Avg.} \\\cline{2-6}
    & S1 & S2 & S3 & S4 & S5 & \\ \hline
    LBP+RR & 13.6 & 58.9 & 37.1 & 21.8 & 23.4 & 31.0\\
    Zhang \etal\cite{zhang2015cross} & 9.8 & \Third{14.1} & 14.3 & 22.2 & \Second{3.7} & 12.9\\
    Zhang \etal \cite{zhang2016single} & {3.4} & 20.6 & \Third{12.9} & 13.0 & 8.1 & 11.6\\
    Sam \etal \cite{sam2017switching} & 4.4 & 15.7 & \Second{10.0} & \Third{11.0} & 5.9 &\Third{9.4}\\
    Liu~\etal\cite{liu2018decidenet}&\Second{2.0}&\Second{13.1}&\First{8.9}&17.4&4.8&\Second{9.3}\\\hline
    L2 &\Third{3.3}&37.9&19.5&\Second{10.5}&\Second{3.7} & 14.9 \\
    SmoothL1 &3.7&45.0&30.1&11.1&\Second{3.7} & 18.7 \\
    Tukey &\Third{3.3}&38.3&19.5&\Second{10.5}&\Second{3.7} & 15.0 \\
    NCL&4.9&14.3&18.7&11.3&\Third{4.6} & 10.7 \\
    \mymodel & \First{1.9} & \First{12.1} & 20.7 & \First{8.3} & \First{2.6} & \First{9.1}\\\hline
  \end{tabular}\label{tab_expo}
\end{table}

\begin{table*}[t]
  \centering
  \renewcommand{\tabcolsep}{8.2pt}
  \caption{Personality prediction bench-marking using mean accuracy $A$
    and coefficient of determination $R^2$ scores.
    The results of the first 6 methods are copied from ~\cite{ponce2016chalearn}
    and ~\cite{zhang2016deep}.
  }\label{tab:personality-comp}
  \begin{tabular}{c|cc|cc|cc|cc|cc|cc} \hline
    \RowsT{}  & \ColsT{|}{Average} & \ColsT{|}{\textit{Extraversion}} &
      \ColsT{|}{\textit{Agreeableness}} & \ColsT{|}{\textit{Conscientiousness}} &
      \ColsT{|}{\textit{Neuroticism}} & \ColsT{}{\textit{Openness}} \\ \cline{2-13}
    & $A$~$\uparrow$   & $R^2$~$\uparrow$ & $A$~$\uparrow$ &
      $R^2$~$\uparrow$ & $A$~$\uparrow$   & $R^2$ ~$\uparrow$ &
      $A$~$\uparrow$   & $R^2$~$\uparrow$  & $A$~$\uparrow$   &
      $R^2$~$\uparrow$  & $A$~$\uparrow$   & $R^2$~$\uparrow$  \\ \hline
    NJU-LAMDA & \Third{0.913} & 0.455 & 0.913 & 0.481 &\Second{0.913} & \Second{0.338} & 0.917 & 0.544 & 0.910 &  0.475 & 0.912 &\Third{0.437} \\
    Evolgen & 0.912 &  0.440 & 0.915 &  0.515 &\Third{0.912} & 0.329 & 0.912 & 0.488 & 0.910 & 0.455 & 0.912 & 0.414 \\
    DCC & 0.911 & 0.411 & 0.911 & 0.431 & 0.910& 0.296 & 0.914 & 0.478 & 0.909 & 0.448 & 0.911 & 0.402 \\
    Ucas & 0.910 & 0.439 & 0.913 & 0.489 & 0.909 & 0.292 & 0.911 & 0.520 & 0.906 & 0.457 & 0.910 & \Second{0.439} \\
    BU-NKU-v1 & 0.909 & 0.394 & 0.916 & 0.514 & 0.907 & 0.234 & 0.913 & 0.487 & 0.902 & 0.363 & 0.908 & 0.372 \\
    BU-NKU-v2 & \Third{0.913} & - & 0.918 &- & 0.907 & - & 0.915 &- & 0.911 &- & \Third{0.914} &- \\
    \hline
    {L2} & \Second{0.915} &\Second{0.467} & \Second{0.920} &\Second{0.544} & \Third{0.912} & \Third{0.333} & \Third{0.918}&0.543 & \Second{0.913} &\Second{0.482} & \First{0.916} &0.426  \\
    {SmoothL1} &\Second{0.915} &\Third{0.466} & \Third{0.919} & \Third{0.542} & \Third{0.912} & 0.332 & \Second{0.919}&\Third{0.548} & \Third{0.912} &\Third{0.480} & 0.913 &0.428 \\
    {Tukey} & \Second{0.915} &\Second{0.467} &  \Third{0.919} &\Third{0.542} & \Third{0.912} & 0.332 & \Second{0.919}&\Second{0.551} & \Third{0.912} & 0.479 & 0.913 &{0.430} \\
    \mymodel & \First{0.918} & \First{0.497} & \First{0.923}&\First{0.571} & \First{0.914}&\First{0.365} & \First{0.922} & \First{0.581} & \First{0.914} & \First{0.512} & \Second{0.915} & \First{0.458} \\ \hline
  \end{tabular}
\end{table*}

\myPara{Shanghaitech dataset} 
The Shanghaitech dataset \cite{zhang2016single} is a large-scale
crowd counting dataset,
which contains 1198 annotated images with a total of 330,165 persons.
This dataset is the largest one in the literature in terms of
the number of annotated pedestrians.
It consists of two parts: Part\_A consisting of 482 images that
are randomly captured from the Internet,
and Part\_B including 716 images that are taken from the busy streets in Shanghai.
Each part is divided into training and testing subset.
The crowd density varies significantly among the subsets,
making it difficult to estimate the number of pedestrians.

We compare our method with six existing methods on the Shanghaitech dataset.
All the detailed results for each method are illustrated in \tabref{tab_tech}.
In the same way, we can see that all deep learning methods outperform
hand-crafted features significantly.
The shallow model in~\cite{zhang2016single} employs a much wider structure
by a multi-column design and performs better than the
shallower CNN models in~\cite{zhang2015cross} in both cases.
A deeper version of the proposed method performs consistently better
than the other shallow one, as expected,
because of employing a much deep pre-trained model.
Moreover, it is interesting to see that with deep negative learning,
even a relatively shallower network structure is on a par with
a much complicated and \sArt switching strategy~\cite{sam2017switching}.
Finally, our deep structure leads to the best performance in terms of
MAE on Part\_A and RMSE on Part\_B.

\myPara{WorldExpo'10 dataset} The WorldExpo'10 dataset \cite{zhang2015cross} is a large-scale and cross-scene
crowd counting dataset.
It contains 1132 annotated sequences which are captured by 108 independent cameras,
all from Shanghai 2010 WorldExpo'10.
This dataset consists of 3980 frames with a total of 199,923 labeled pedestrians,
which are annotated at the centers of their heads.
Five different regions of interest (ROI) and the perspective maps are
provided for the test scenes.

We follow the standard evaluation protocol and use all the training frames
to learn our model.
For comparison, the quantitative results are given in \tabref{tab_expo}.
In the same way, we observe that learned representations are more robust
than the handcraft features.
Even without using the perspective information,
our results are still comparable with another deep learning method
\cite{zhang2015cross} which used perspective normalization to crop
$3\times3$ square meters patches with 0.5 overlaps on testing time.
The deeper version of our proposed method outperforms all other
in terms of average performance.

\subsection{Personality~Analysis}

For personality analysis, the ensemble size is set to be 16. We use the ChaLearn personality dataset~\cite{ponce2016chalearn},
which consists of $10k$ short video clips with 41.6 hours (4.5M frames) in total.
In this dataset, people face and speak to the camera.
Each video is annotated with personality attributes as the Big Five
personality traits (\emph{Extraversion, Agreeableness, Conscientiousness,
Neuroticism and Openness})~\cite{ponce2016chalearn} in $[0,\,1]$.
The annotation was done via Amazon Mechanical Turk.
For the evaluation, we follow the standard protocol in
ECCV 2016 ChaLearn First Impression Challenge~\cite{ponce2016chalearn},
and use the mean accuracy $A$ and coefficient of determination $R^2$,
which are defined as follows:
\begin{equation}\label{eq:mean_accuracy}
	A = 1 - \frac{1}{N^t} \sum_{i}^{N^t} |\Ymat^P_i - \mathbf{P}_i|,
\end{equation}
\begin{equation}\label{eq:Rtwo}
	R^2 = 1 - \sum_{i}^{N^t} (\Ymat^P_i - \mathbf{P}_i)^2/
              \sum_{i}^{N^t} (\mathbf{\bar{Y}}^P -\mathbf{P}_i)^2,
\end{equation}
where $N^t$ denotes the total number of testing samples,
$\Ymat^P$ the ground truth, $\mathbf{P}_i$ the prediction,
and $\mathbf{\bar{Y}}^P$ the average value of the ground truth.

\begin{table}[ht]
  \centering \renewcommand{\tabcolsep}{9.5pt}
  \caption{Comparison of the properties of the proposed method vs.
    the top teams in the 2016 ChaLearn First Impressions Challenge.
  }\label{tab:chalearn-properties}
  \begin{tabular}{c|c|c|c|c} \hline
    & \RowsT{Fusion} & \ColsT{|}{Modality} & \RowsT{End-to-End} \\\cline{3-4}
    & & Audio& Video & \\\hline
    Ours          & late & \xmark & \cmark & \cmark \\\hline
    NJU-LAMDA$^1$ & late & \cmark & \cmark & \cmark \\\hline
    Evolgen       & early& \cmark & \cmark & \cmark \\\hline
    DCC           & late & \cmark & \cmark & \cmark \\\hline
    Ucas          & late & \cmark & \cmark & \xmark \\\hline
    BU-NKU-v1     & early& \xmark & \cmark & \xmark \\\hline
    BU-NKU-v2$^2$ & early& \cmark & \cmark & \xmark \\\hline
  \end{tabular}
  \vspace{-0.7em}\\
  \bigskip {$^1$ winner, $1^{st}$ ChaLearn First Impressions Challenge (ECCV 2016). \\
  $^2$ winner, $2^{nd}$ ChaLearn First Impressions Challenge (ICPR 2016)}
\end{table}

% We utilize a truncated 20 layer version of the \emph{SphereFace}
% model~\cite{liu2017sphereface} for personality analysis.
% We first detect and align faces for each input
% with well-established \emph{MTCNN}~\cite{zhang2016joint}.
% In order to speed up training and reduce the risk of over-fitting,
% we take a similar approach as done in~\cite{wang2016temporal}
% to first sparsely sample 10 frames from each video in a randomized manner.
% Average pooling is further used to aggregate multiple results
% for the same video.
We train the whole network with an initial learning rate of $0.01$.
For each mini-batch, we randomly select 10 videos thus generating
a total batch size of 100.
We set $\nTrees=8$ in this experiment and train the network for
$28k$ iterations and decrease the learning rate by a factor of 10
in the $16k^{th}$, $24k^{th}$ and $28k^{th}$ iteration.

The quantitative comparison between the proposed method and other \sArt
works on personality recognition is shown in \tabref{tab:personality-comp}.
Moreover, \tabref{tab:chalearn-properties} lists the comparison of
the details of several latest personality recognition methods.
In contrast to other approaches,
ours can be trained end-to-end using only one pre-trained model.
Moreover, unlike most methods which fuse both acoustic and visual cues,
our proposed method uses only video frames as input.
The teams from NJU-LAMDA to BU-NKU-v1 are the top five participants
in the $1^{st}$ ChaLearn Challenge on First Impressions~\cite{ponce2016chalearn}.
Note that BU-NKU was the only team not using audio in the challenge,
and their predictions were rather poor comparatively.
After adding the acoustic cues,
the same team won the $2^{nd}$ ChaLearn Challenge on First
Impressions~\cite{ponce2016chalearn}.
Importantly, our methods only consider visual streams.
Firstly, we observe that the deeply learned representations
are well transferable between face verification and personality analysis.
This can be verified by the last four results in \tabref{tab:personality-comp}.
By utilizing \sArt face verification network and good practices
in video classification~\cite{wang2016temporal},
those methods outperform current state-of-the-arts.
Secondly, L2 and SmoothL1 loss and Tukey Loss all lead to
comparably good results for this task.
Finally, the proposed method outperforms all the methods on both metrics
in all scenarios.

\newcommand{\HumanWks}{Human workers \cite{han2015demographic}}
\newcommand{\AGES}{AGES \cite{geng2007automatic}}
\newcommand{\MTWGP}{MTWGP \cite{zhang2010multi}}
\newcommand{\CASVR}{CA-SVR \cite{chen2013cumulative}}
\newcommand{\SVR}{SVR \cite{guo2008image}}
\newcommand{\OHRank}{OHRank \cite{chang2011ordinal}}
\newcommand{\DLA}{DLA \cite{wang2015deeply}}
\newcommand{\Rank}{Rank \cite{chang2010ranking}}
\newcommand{\Rothe}{Rothe et al. \cite{rothe2016some}}
\newcommand{\DEX}{DEX \cite{rothe2018deep}}
\newcommand{\dLDLF}{dLDLF~\cite{shen2017label}}
\newcommand{\ARN}{ARN~\cite{agustsson2017anchored}}
\newcommand{\DRF}{DRF ~\cite{Shen_2018_CVPR}}
\newcommand{\DIF}{DIF \cite{han2015demographic}}
\newcommand{\CPNN}{CPNN \cite{geng2013facial}}
\newcommand{\CAM}{CAM~\cite{luu2011contourlet}}

\begin{table}[thb]
\centering \renewcommand{\tabcolsep}{9.5pt}
\caption{Results of different age estimation methods on the
  MORPH~\cite{ricanek2006morph} and FG-NET~\cite{panis2016overview} datasets.}
\begin{tabular}{c|c|c|c|c} \hline
    \RowsT{Method} & \ColsT{|}{MORPH} & \ColsT{}{FG-NET} \\  \cline{2-5}
    & MAE~$\downarrow$ & CS~$\uparrow$ & MAE~$\downarrow$ & CS~$\uparrow$ \\ \hline
    \HumanWks& 6.3& 51.0 & 4.70 & 69.5 \\
    \AGES  & 8.83 & 46.8 & 6.77 & 64.1 \\
    \MTWGP & 6.28 & 52.1 & 4.83 & 72.3 \\
    \CASVR & 5.88 & 57.9 & 4.67 & 74.5 \\
    \SVR   & 5.77 & 57.1\\
    LARR \cite{guo2008image}&&& 5.07 & 68.9 \\
    \OHRank& 6.07 & 56.3 & 4.48 & 74.4 \\
    \DLA   & 4.77 & 63.4 & 4.26 &  -   \\
    \Rank  & 6.49 & 49.1 & 5.79 & 66.5 \\
    \DIF   &  -   &  -   & 4.80 & 74.3 \\
    \CPNN  &  -   &  -   & 4.76 &  -   \\
    \CAM~  &  -   &  -   & 4.12 &  -   \\
    \Rothe & 3.45 &  -   & 5.01 &  -   \\
    \DEX   & 3.25 &  -   & 4.63 &  -   \\
    \dLDLF & 3.02 & 81.3 &  -   &   -  \\
    \ARN   & 3.00 &  -   &  -   &   -  \\
    \DRF   &\Second{2.91}&\Third{82.9} & \Second{3.85} & 80.6\\
    \hline
    SmoothL1 & 2.99 & 82.5 & 3.95 &\Third{80.9} \\
    Tukey&\Third{2.90}&\Second{83.1}&\Third{3.87}&\First{82.5}\\
    \mymodel & \First{2.85} &\First{83.8}&\First{3.71}&\Second{81.8}\\\hline
\end{tabular}\label{tab-morph}
\end{table}

\renewcommand{\addImg}[3]{\subfloat[#1]{\includegraphics[width=0.33\linewidth]{image/#2.png}\label{fig:sup:#3}}}

\begin{figure*}[!thb]
\centering
\begin{minipage}[]{0.33\linewidth} \centering
  \subfloat[Original Image]{\includegraphics[width=\linewidth]{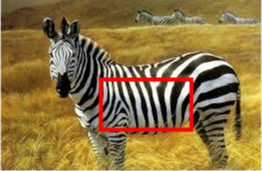}\label{fig:sup:original}}
  \label{fig:suo:original}
\end{minipage}
\begin{minipage}[]{0.66\linewidth} \centering
  \addImg{HR GT (PSNR, SSIM)}{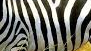}{gt} \hfill
  \addImg{Bibubic (12.8, 18.7)}{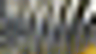}{bic} \hfill
  \addImg{DRRN (13.3, 30.2)}{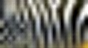}{drrn} \\
  \addImg{SmoothL1~(13.3, 29.5)}{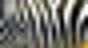}{l1} \hfill
  \addImg{Tukey (13.3, 30.2)}{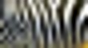}{tukey} \hfill
  \addImg{Ours (13.4, 30.4)}{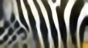}{ours}
\end{minipage}
\caption{Visual comparison for $4\times$ super-resolution of different
  super-resolution results.
  \figref{fig:sup:gt} shows the ground-truth high resolution image cropped
  from the original image in \figref{fig:sup:original}.
}\label{fig:super}
\end{figure*}

\subsection{Age Estimation}
\label{sec:age_exp}

\begin{table*}[!thb]
\caption{Average PSNR/SSIM/IFC score for image super-resolution of scale factor
  $\times2$, $\times3$ and $\times4$ on datasets Set5, Set14, BSD100 and Urban100.}
\centering  \renewcommand{\tabcolsep}{2.1pt}
\begin{tabular}{l|l|l|l|l|l|l|l|l|l|l}\hline
  \ColsT{|}{\RowsT{Dataset}} & \multicolumn{9}{c}{PSNR/SSIM/IFC~$\uparrow$}\\\cline{3-11}
  \ColsT{|}{} &SRCNN~\cite{dong2016image} &SelfEx~\cite{huang2015single} &RFL~\cite{schulter2015fast}&VDSR~\cite{kim2016accurate}&DSRN~\cite{han2018image}&DRRN~\cite{tai2017image}  &Tukey &SmoothL1& \mymodel\\\hline
\multirow{3}{*}{set5}
  & $\times2$ & 36.7/95.4/8.04 & 36.5/95.4/7.81 & 36.5/95.4/8.56 & {37.5}/{95.9}/\Third{8.57} &
    \Third{37.7}/\Third{95.9}/8.6 & \Second{37.7}/\Second{95.9}/\Second{8.67} &
    {37.5}/\First{95.9}/7.58 & 37.0/95.6/7.08 & \First{37.8}/\First{95.9}/\First{8.70} \\
  & $\times3$ & 32.8/90.9/4.66 & 32.6/90.9/4.75 & 32.4/90.6/4.93 & 33.7/92.1/\Third{5.22} &
    \Third{33.9}/\Third{92.2}/\Third{5.22} & \Second{34.0}/\Second{92.4}/\Second{5.49} &
      33.1/91.6/4.65 & 33.0/91.6/4.58 & \First{34.1}/\First{92.5}/\First{5.43} \\
  & $\times4$ & 30.5/86.3/2.99 & 30.3/86.2/3.17 & 30.1/85.5/3.19 &
      31.6/\Third{88.4}/\Third{3.55} & \Third{31.4}/88.3/3.50 & \Second{31.7}/\Second{88.9}/\Second{3.70} &
      30.8/87.4/2.78 & 30.4/87.1/2.73 & \First{31.7}/\First{89.0}/\First{3.73} \\\hline
\multirow{3}{*}{set14}
  & $\times2$ & 32.5/90.7/7.78 & 32.2/90.3/7.59 & 32.3/90.4/8.18 & \Third{33.0}/91.2/\Third{8.18} &
    \Second{33.2}/\Third{91.3}/8.17 & \First{33.2}/\Second{91.4}/\First{8.32} &
    32.7/86.9/7.48 & 32.2/87.0/7.10 & \First{33.2}/\First{91.4}/\Second{8.30} \\
  & $\times3$ & 29.3/82.2/4.34 & 29.2/82.0/4.37 & 29.1/81.6/4.53 & 29.8/83.1/4.73 &
    \Second{30.3}/\Second{83.7}/\First{4.89} & \Third{30.0}/\Third{83.5}/\Third{4.88} &
    29.4/82.7/4.22 & 29.5/82.4/4.17 & \First{30.0}/\First{83.5}/\Second{4.89} \\
  & $\times4$ & 27.5/75.1/2.75 & 27.4/75.2/2.89 & 27.2/74.5/2.92 & 28.0/76.7/3.13 &
    \Third{28.1}/\Third{77.0}/\Third{3.15} & \Second{28.2}/\Second{77.2}/\Second{3.25}  &
    27.7/76.0/2.87 & 27.1/75.4/2.84 & \First{28.3}/\First{77.3}/\First{3.28} \\\hline
\multirow{3}{*}{BSD100}
  & $\times2$ & 31.4/0.89/- & 31.2/88.6/- & 31.2/88.4/- & 31.9/89.6/- &
    \First{32.1}/\Third{89.7}/- & \Third{32.1}/\Second{89.7}/- & 31.7/88.4/- &
    31.6/88.3/- & \Second{32.1}/\First{89.8}/- \\
  & $\times3$ & 28.4/78.6/- & 28.3/78.4/- & 28.2/78.1/- &
    \Third{28.8}/\Third{79.8}/- & 28.8/79.7/- & \Second{29.0}/\Second{80.0}/- &
    28.6/79.3/- & 28.2/78.8/- & \First{29.0}/\First{80.1}/- \\
  & $\times4$ & 26.9/71.0/- & 26.8/71.1/- & 26.8/70.5/- &
    \Third{27.3}/\Third{72.5}/- & 27.3/72.4/- & \Second{27.4}/\Second{72.8}/- &
    27.0/71.8/- & 27.0/70.8/- & \First{27.4}/\First{72.9}/-\\\hline
\multirow{3}{*}{Urban100}
  & $\times2$ & 29.4/89.5/7.99 & 29.5/89.7/7.94 & 29.1/89.0/8.45 & 30.8/91.4/\Third{8.65} &
    \Third{31.0}/\Third{91.6}/8.60 & \Second{31.2}/\Second{91.9}/\Second{8.92} &
    30.3/90.9/7.87 & 30.1/90.1/7.25 & \First{31.3}/\First{92.0}/\First{8.95}\\
  & $\times3$ & 26.2/79.9/4.58 & 26.4/80.9/4.84 & 25.9/79.0/4.80 & 27.1/82.8/\Third{5.19} &
    \Third{27.2}/\Third{82.8}/5.17 & \Second{27.5}/\Second{83.8}/\First{5.46} &
    26.6/81.4/4.58 & 26.5/81.4/4.52 & \First{27.6}/\First{83.8}/\Second{5.47}\\
  & $\times4$ &24.5/72.2/2.96 & 24.8/73.7/3.31 & 24.2/71.0/3.11 &
    \Third{25.2}/\Third{75.2}/\Third{3.50} & 25.1/74.7/3.30 & \Second{25.4}/\Second{76.4}/\Second{3.68} &
    24.7/73.3/2.93 & 24.1/73.2/2.86 & \First{25.6}/\First{76.5}/\First{3.71}\\\hline
\end{tabular}
\label{tab:super}
\end{table*}

% For Age estimation,
% we use the network backbone of deep forest~\cite{Shen_2018_CVPR}.
% It reformulates the split node of a decision forest as a fully connected layer
% of a CNN and learns both split node and lead node in an iterative manner.
% More specifically, by fixing the leaf nodes, the split nodes as well as the CNN parameters are optimized by Back-propagation.
% Then, by fixing the split nodes, the leaf nodes are optimized by
% iterating a step-size free and fast converging update rule
% derived from Variational Bounding.
% Instead of using this iterative strategy,
% we use the proposed NCL loss in each node to make them both accurate
% and diversified.

We use the same training and evaluation protocol as done in~\cite{Shen_2018_CVPR}.
More specifically, we first use a standard face detector to detect faces
~\cite{viola2001rapid} and further localized the facial landmarks
by AAM~\cite{cootes2001active}. The ensemble size is 5.
After that, we perform face alignment to guarantee all eyeballs stay
at the same position in the image.
We further augment the training data by the following strategies:
(1) cropping images with some random offsets,
(2) adding Gaussian noise to the original images,
and (3) randomly flipping from left to right.
We compare the proposed method with various state-of-the-arts
on two standard benchmarks:
MORPH~\cite{ricanek2006morph} and FG-NET~\cite{panis2016overview}.

As for the evaluation metric,
we follow the existing method and choose
\textit{Mean Absolute Error} (MAE) as well as \textit{Cumulative Score} (CS).
CS is calculated by $\textit{\text{CS}}(l) = \frac{N_l}{N}·100\%$,
where $N$ is the total number of testing images and $N_l$ is
the number of testing facial images whose absolute
error between the estimated age and the ground truth age is
not greater than $l$ years.
Here, we set the same error level 5 as in~\cite{Shen_2018_CVPR}.

We first summarize our results on MPRPH dataset in \tabref{tab-morph}.
It contains more than 55,000 images from about 13,000 people of different races.
We perform our evaluation on the first setting~(setting I)~\cite{Shen_2018_CVPR}
which selects 5,492 images of Caucasian Descent people from the original
MORPH dataset to reduce the cross-ethnicity effects.
In this setting, these 5,492 images are randomly partitioned into two subsets:
$80\%$ of the images are selected for training and others
for testing. The random partition is repeated 5 times, and
the final performance is averaged over these 5 different partitions.
Since the DRF method~\cite{Shen_2018_CVPR} assumed each leaf node was
a normal distribution,
minimizing negative log likelihood loss was equivalent to minimize
the L2 loss of each node~\footnote{Actually the released implementation
of DRF~\cite{Shen_2018_CVPR} used L2 loss to avoid observing negative
loss during training.}.
As can be seen from \tabref{tab-morph},
the proposed method achieves the best performance on this dataset,
and outperforms the current state-of-the-arts with a clear margin.

We then conduct experiments on FG-NET~\cite{panis2016overview},
which contains 1002 facial images of 82 individuals.
Each individual in FG-NET has more than 10 photos taken at different ages.
The FG-NEt data is challenging because each image may have
a large variation in lighting conditions, poses and expressions.
We follow the protocol of~\cite{Shen_2018_CVPR} to perform ``leave one out"
cross validation on this dataset.
The quantitative comparisons on FG-NET dataset are shown in \tabref{tab-morph}.
As can be seen, our method achieves better results
(MSE: 3.71 vs 3.85 and CS: 81.8 vs 80.6) than DRF ~\cite{Shen_2018_CVPR}.

\subsection{Image Super-resolution}

% For image super-resolution, we choose the \sArt DRRN~\cite{tai2017image}
% as our network backbone and change the L2 loss into the proposed NCL loss.
%
We follow exactly the same training and evaluation protocol.
More specifically, by following~\cite{kim2016accurate,schulter2015fast},
a training dataset of 291 images,
where 91 images are from Yang et al. ~\cite{yang2010image} and other
200 images are from Berkeley Segmentation Dataset~\cite{martin2001database},
were utilized for training.
Finally, the method was evaluated on Set5~\cite{bevilacqua2012low},
Set14~\cite{zeyde2010single}, BSD100~\cite{martin2001database} and
Urban100~\cite{huang2015single} dataset,
which have 5, 14, 100 and 100 images respectively.
The initial learning rate is set to 0.1 and then decreased to
half every 10 epochs.
Since a large learning rate is used in our work,
we adopt the adjustable gradient cliping~\cite{tai2017image}
to boost the convergence rate while suppressing exploding gradients.
Peak Signal-to-Noise Ratio (PSNR),
Structural SIMilarity (SSIM)~\cite{wang2004image} and
Information Fidelity Criterion (IFC)~\cite{sheikh2005information}
were used for the quantitative evaluations.

\tabref{tab:super} summarizes the main results of both PSNR in db and
SSIM ($\times 100\%$) on the four testing sets.
Similarly, the results of IFC are presented in \tabref{tab:super}.
Firstly, we can find that the image super-resolution is extremely challenging
as most state-of-the-art approaches perform comparably well.
However, it is still obvious that the proposed method outperforms the original
L2 loss in most cases,
leading to even better results than a more recent work using
dual-state recurrent networks~\cite{han2018image}.
In addition, other loss functions such as SmoothL1 and Tukey loss are both
outperformed by L2 loss in a large margin.
Qualitative comparisons among DRRN~\cite{tai2017image} and SmoothL1,
Tukey and our proposed method are illustrated in \figref{fig:super}.
As we can see, our method produces relatively sharper edges with respect to patterns,
while other methods may give blurry results.

\subsection{Discussions}

After demonstrating the superiority of the proposed method by extensively
comparing them with many \sArt methods on multiple datasets,
we now provide more discussions to shed light upon their rationale and
sensitivities with some hyper-parameters.

\begin{table}[t]
\caption{Comparing the performance of NCL and conventional ensemble on
  the crowd counting task.}
\centering \renewcommand{\tabcolsep}{7pt}
\begin{tabular}{c|c|c|c|c} \hline
  \RowsT{Datasets}& \ColsT{|}{Conventional Ensemble} & \ColsT{}{\mymodel} \\ \cline{2-5}
  & MAE~$\downarrow$ & RMSE~$\downarrow$ & MAE~$\downarrow$ & RMSE~$\downarrow$\\ \hline
  UCF\_CC\_50          & 380.5 & 527.2& \First{288.4}& \First{404.7} \\ \hline
  Shanghaitech Part\_A & 91.6  & 127.9& \First{73.5} & \First{112.3} \\ \hline
  Shanghaitech Part\_B & 21.3  & 30.9 & \First{18.7} & \First{26.0} \\ \hline
  WorldExpo'10         & 16.4  & -    & \First{9.1}  & - \\ \hline
\end{tabular}\label{tab_ensamble}
\end{table}

\myPara{NCL or Conventional Ensemble Learning?}In \tabref{tab_ensamble}, we compared the performance of the proposed method
with conventional ensemble learning and choose crowd counting as a study case.
It is widely accepted that training deep networks like VGG remains to be challenging.
In~\cite{boominathan2016crowdnet}, a shallow network was proposed to assist the
training and improve the performance of deep VGG network.
When compared with results achieved on dataset UCF\_CC\_50 by other methods
shown in \tabref{tab_ucf},
our implementation of a conventional ensemble method using a single VGG network
leads to much improved results.
However, it still over-fits severely compared with other \sArt methods.
More specifically, it was outperformed by recent methods such as multi-column
structure~\cite{zhang2016single}, multi-scale Hydra method~\cite{onoro2016towards},
and advanced switching strategy~\cite{sam2017switching}.
In contrast, the proposed method leads to much improved performance compared
with this baseline in all cases and outperforms all the aforementioned methods.
As illustrated in \figref{steady_state},
the NCL mechanism used here encourages diversities in the ensemble and
thus it is more likely to allow error canceling.  \zl{For more results on other datasets, please refer to~\tabref{r1-ensemble} and ~\tabref{r2-ensemble}.}

The learning objective function in \equref{NCL} is also in line with
Breiman's strength-correlation theory~\cite{breiman2001random}
on the VC-type bound for the generalization ability of ensemble models,
which advocated both accurate and decorrelated individual models.
It as well appreciated that the individual model should be able to
exhibit different patterns of generalization--
a very simple intuitive explanation is that a million identical estimators
are obviously no better than a single.

\myPara{DNCL for other loss functions} \zl{It is widely-accepted that encouraging the diversity could generate better ensemble. Although DNCL is derived under the commonly used L2 loss function, here we show that naively apply this idea to other loss functions could be beneficial. To this end, we replace the first part in Eqn.~\ref{NCL} with other loss functions while keep the second part unchanged to make the ensemble negatively-correlated. We report the  detailed results in ~\tabref{r1-ensemble} and ~\tabref{r2-ensemble}. Firstly, the results show 
that the proposed ensemble strategy still generates better results  than single model for each loss function but is outperformed by the proposed method. Secondly, one can observe that in some cases NCL with other loss functions were outperformed by conventional ensemble. This indicates that another diversity measurements~\cite{tang2006analysis} could be better when other loss functions were utilized.  }

\begin{table}
\caption{\zl{Results of different ensemble strategies on the age and crowd datasets.}}\label{r1-ensemble}
\centering
\begin{tabular}{c|c|c|c|c|c|c}\hline
 Dataset& \multicolumn{2}{c|}{MORPH} &  \multicolumn{2}{c}{ShanghaiA} \\\hline
 Metric&MAE$\downarrow$  &   CS$\uparrow$ &MAE$\downarrow$&RMSE$\downarrow$ \\\hline
Proposed& \First{2.35} &  \First{83.8} & \First{73.5} & \First{112.3} \\\hline
L1-NCL& 2.94 &  82.8  &77.9&118.86\\\hline
Tukey-NCL&2.87&83.4&86.3&121.7\\\hline
 Conventional Ensemble& 2.89&83.1&91.6&127.9\\\hline
\end{tabular}
\end{table}

\begin{table}
  \centering
  \renewcommand{\tabcolsep}{8.2pt}
  \caption{\zl{Results of different ensemble strategies on the personality and the Urban100 (scale factor
  of $\times4$) dataset.}}\label{r2-ensemble}
  \resizebox{0.96\linewidth}{!}{%
  \begin{tabular}{c|cc|c} \hline
    \RowsT{}  & \ColsT{|}{Chalearn (Average Score)}   & {Urban100 ($4\times4$)} \\ \cline{2-4}
    & $A$~$\uparrow$   & $R^2$~$\uparrow$&  PSNR/SSIM/IFC~$\uparrow$\\ \hline
    % {L2} & 0.915 & 0.467 & 0.920 & 0.544 & 0.912 & 0.333 & 0.918 & 0.543 & 0.913 & 0.482 & 0.916 & 0.426  \\
    % {SmoothL1} & 0.915 & 0.466 & 0.919 & 0.542 & 0.912 & 0.332 & 0.919 & 0.548 & 0.912 & 0.480 & 0.913 & 0.428 \\
    Proposed & \First{0.918} & \First{0.497}&\First{\First{25.6}}/\First{76.5}/\First{3.71} \\\hline
    {L1-NCL} & 0.916 & 0.468 & 24.6/75.2/3.46 \\\hline
    % {Tukey} & 0.915 & 0.467 & 0.919 & 0.542 & 0.912 & 0.332 & 0.919 & 0.551 & 0.912 & 0.479 & 0.913 & 0.430 \\
    {Tukey-NCL} & 0.915 & 0.467 &24.9/74.1/3.21 \\\hline
    {Conventional Ensemble } & 0.917 & 0.470&24.7/75.0/3.44 \\\hline
  \end{tabular}
  }
\end{table}

% \begin{table}
% \caption{Results of different ensemble strategies on the Urban100 dataset (scale factor
%   of $\times4$)} \label{r3}
% \centering  
% \begin{tabular}{l|l|l}\hline
%   \multicolumn{3}{c}{PSNR/SSIM/IFC~$\uparrow$}\\\hline
%   Tukey-NCL &L1-NCL& Proposed\\\hline
% 24.9/74.1/3.21 & 24.6/75.2/3.46 & \First{25.6}/\First{76.5}/\First{3.71}\\\hline
% \end{tabular}
% \end{table}

\myPara{Effect of \texorpdfstring{$\lambda$}{lambda} and
\texorpdfstring{$\nTrees$}{K}.}Parameter $\lambda$ controls the correlation between each model in the ensemble.
On the one hand, setting $\lambda=0$ is equivalent to train each regressor
in an independent manner.
On the other hand, employing a larger value for $\lambda$ overemphasizes
the effect of diversity and may lead to poor individual regressors.
We empirically find that setting $\lambda$ to be a relatively smaller value
$\lambda \in [10^{-3},10^{-2}]$ usually leads to satisfactory results.
Parameter $\nTrees$ stands for the number of base regressors in the ensemble.
Theoretically speaking, conventional ensemble learning such as bagging
and decision tree ensemble requires larger ensemble
sizes~\cite{rodriguez2006rotation, fernandez2014we,zhang2017benchmarking}
to perform well.
However, with the constraint of using the same amount of parameter,
increasing the value of $\nTrees$ will pass each base model less input information,
which may lead to worse performance.
We empirically find that the proposed method works well even with a relatively
smaller ensemble size.
For crowd counting,  setting $\nTrees$  to be within 32 and 64 can generate
satisfactory results and it is set to be 64 by default as no
significant improvement is observed with a more number of regressors. \zl{More detailed report on the effect of $\nTrees$ are provided in Table~\ref{k}.} Similarly, the performances of personality analysis and image super-resolution
are stable when $\nTrees$ is within [8,16] and [16,32] and they are set to be
8 and 16, respectively.
For age estimation, we use the same ensemble size of 5 as done in the
original paper~\cite{Shen_2018_CVPR}.

\begin{figure}[!tbh]
  \centering
  \includegraphics[width=0.35\linewidth]{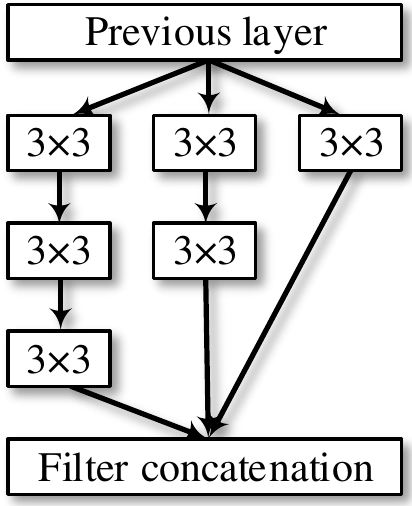}
  \caption{The Multi-Scale Blob module used in NCL.}
  \label{fig:MSB}
\end{figure}

\begin{table}
    \centering
    \caption{\zl{Effect of $\nTrees$ on Shanghai Part A dataset.}}
    \begin{tabular}{c|c|c|c|c|c|c|c}\hline
      $\nTrees$  &1&16&32&64&128&256&512  \\\hline
        MAE$\downarrow$ &105.4&94.1&79.1&\First{73.5}&179.4&433.3&433.5\\\hline 
        RMSE$\downarrow$&152.3&138.8&113.1&\First{112.3}&257.1&560.2&560.2\\\hline
    \end{tabular}
    
    \label{k}
\end{table}
\myPara{Independent of the network backbone.} \label{sec:network}While tremendous  progress  has  been  achieved in vision community by
aggressively exploring deeper ~\cite{he2016deep} or wider
architectures~\cite{Zagoruyko2016WRN},
specially-designed network architecture~\cite{shi2016rank,liu_tpami_2018},
or heuristic engineering tricks~\cite{he_bag} with the
standard “convolution + pooling” recipe,
we want to emphasize that the proposed method is independent of the network
backbone and almost complementary to those strategies.
To show this , we first observe that combing the proposed NCL
learning strategies with each ``special-purpose" network in each task can
lead to improved results.
In order to further demonstrate the independence between the proposed method
and the network backbone,
we choose crowd counting as an example and train a relatively shallower model
named as NCL,
which is constructed by stacking several Multi-Scale Blob as shown in
\figref{fig:MSB},
aiming to increase the depth and expand the width of the crowd model in a single network.
Multi-Scale Blob (MSB) is an Inception-like model which enhances
the feature diversity by combining feature maps from different network branches.
More specifically, it contains multiple filters with different kernel size
(including $7\times 7$, $5\times5 $ and $3\times3 $).
This also makes the net more sensitive to crowd scale changing of the images.

Motivated by VGGNet~\cite{simonyan2014very}, to make the model more discriminative,
we further achieve  $5\times5$ and $7\times7$ convolutional layers by stacking
two and three $3\times3$ convolutional layers, respectively.
In our adopted network, the first convolution layer consists of
$16 \times 5\times 5$ filters and is followed by a $2\times2$ max pooling layer.
After that, we stack two MSB modules as demonstrated in \figref{fig:MSB},
where the first MSB module is followed by a $2\times2$ max-pooling layer.
The number of feature maps of each convolution layer in these two MSB modules
is 24 and 32, respectively.
Finally, we use the same $1\times 1$ convolution layer on each of the feature
maps as regressor $\mapping^\nStage_k$ to get the final crowd density map.

The main results of the shallow network can be found in \tabref{tab_ucf},
\tabref{tab_tech} and \tabref{tab_expo}.
With the proposed negative correlation learning strategy,
it is also interesting to see that
1)both our deep and shallow networks work well;
2) deep networks (\mymodel) are better than shallower networks (NCL),
as expected.
However, the shallower network (NCL) still leads to competitive results and
may be advantageous in resource-constrained scenarios as
it is computationally cheaper.

% As another example, we also test NCL on age estimation with another network backbone. More specifically, we choose the network backbone from DEX~\cite{rothe2018deep} which has been reported to be worse than deep forest~\cite{Shen_2018_CVPR}. Comparing with DEX which achieves an MSE of 3.25 on Morph dataset~\cite{ricanek2006morph}, the proposed method leads to a better MSE　of.

\myPara{Other Aggregation Methods} \zl{Our loss function is derived under the widely-used ensemble setting in which each base model is assigned with equal importance. In this part we investigate the effect of DNCL when different base model has different importance. To this end, we use another $1 \times 1$ convolution to aggregate the results from each base model and report the results the on Shanghaitech Part A dataset and the
results are summarized in  ~\tabref{other-aggre}. The experimental results shows that
the proposed methods achieve better results. However, as the diversities are also
enhanced in the ``weighted average" method, the results are better than conventional
ensemble, as expected.}
\begin{table}[!thb]
  \centering
  \caption{\zl{Comparison of different aggregation methods on the Shanghaitech Part A dataset.}}
   \label{other-aggre}
  \begin{tabular}{c|c|c} \hline
                          & MAE  & RMSE         \\\hline
    Conventional Ensemble & 91.6 & 127.9        \\\hline
    DNCL           &\First{73.5} & \First{112.3}\\\hline
    Weighted average       & 83.5 & 120.8        \\\hline
  \end{tabular}
  \end{table}

\myPara{Visualization of the Enhanced Diversities.} In this section we provide another evidence to show the enhanced diversities
in our ensemble methods.
We choose crowd counting as our studying case and compute
their pair-wise Euclidean distance between each pair of the predictions
from each base model.
From \figref{fig:dmap:enseD} and \figref{fig:dmap:nclD},
we can easily observe that there exist a larger discrepancy in the proposed method,
as we expected.
Finally, a more diversified ensemble leads to better final performance,
as can be found in \figref{fig:dmap:gt}, \ref{fig:dmap:ense} and ~\ref{fig:dmap:ncl}.

\renewcommand{\addImg}[3]{\subfloat[#1]{\includegraphics[width=0.47\linewidth,height=0.33\linewidth]{#2.jpg}\label{fig:dmap:#3}}}
\newcommand{\addImgW}[3]{\subfloat[#1]{\includegraphics[width=0.47\linewidth,height=0.47\linewidth]{#2.jpg}\label{fig:dmap:#3}}}

\begin{figure}[!thb]
  \centering \renewcommand{\tabcolsep}{1pt} \small
  \begin{tabular*}{\linewidth}{ccc}  &
  \addImg{Original Image}{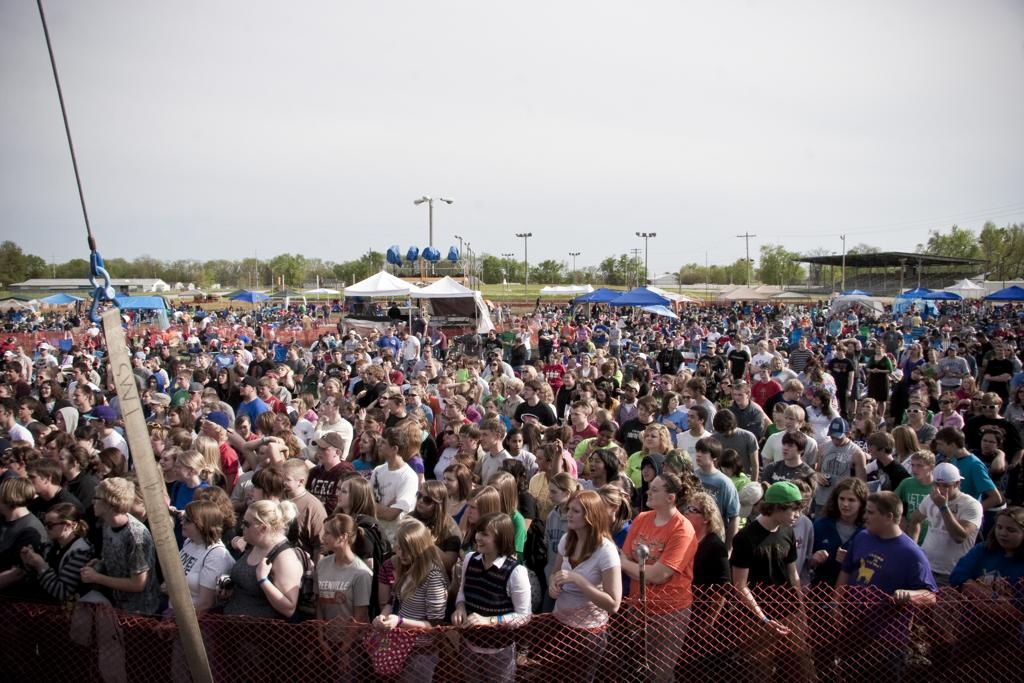}{ori} &
  \addImg{Density Map (568)}{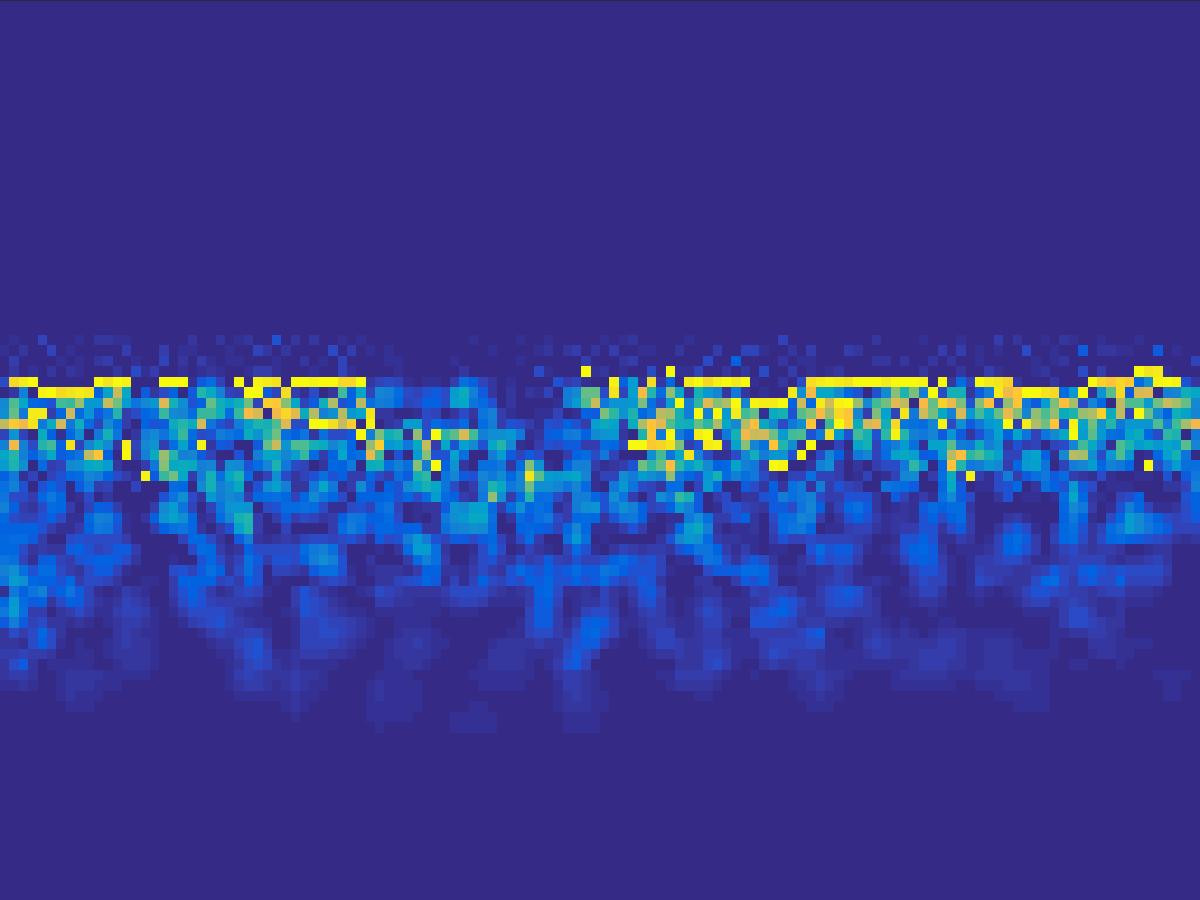}{gt} \vspace{-.1in} \\
  \begin{sideways} ~~~ Prediction results \end{sideways}&
  \addImg{Conventional ensemble (679)}{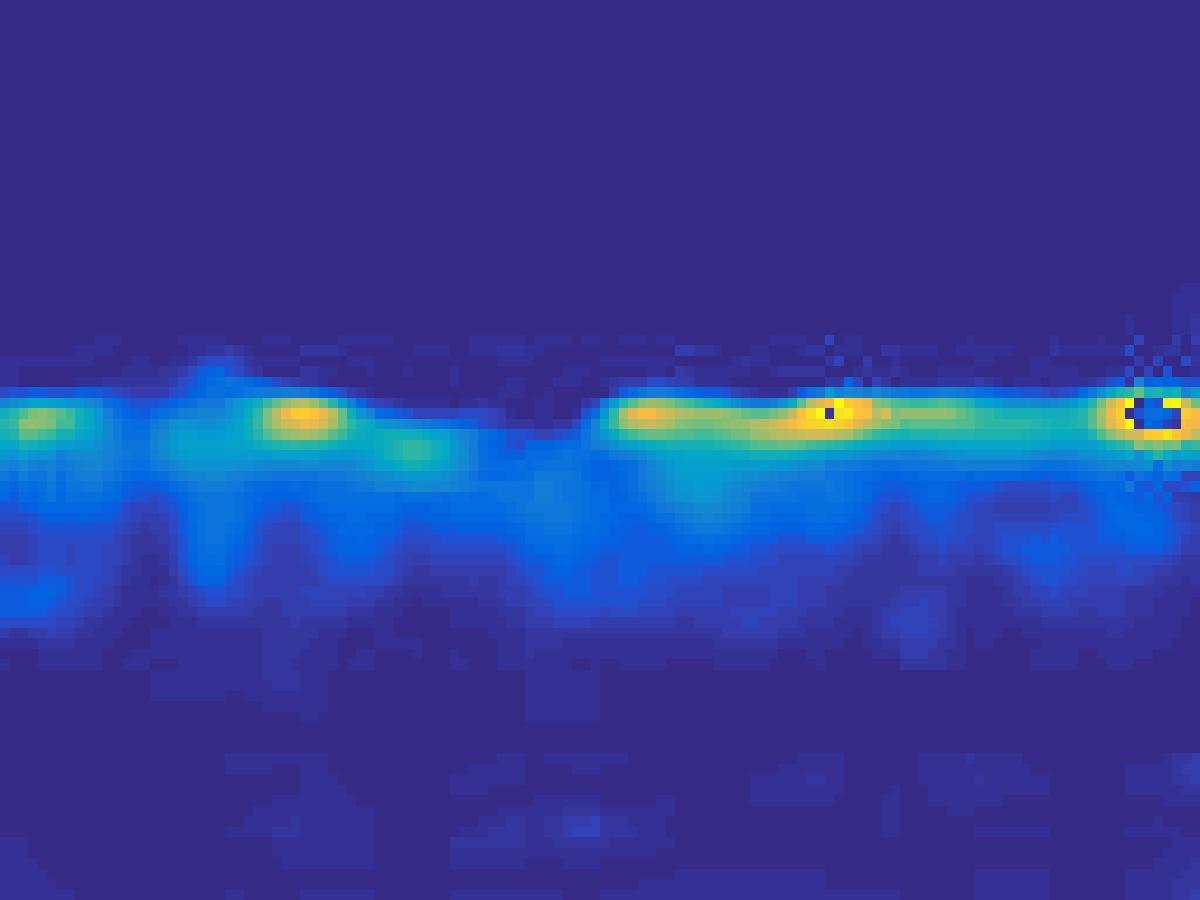}{ense} &
  \addImg{DNCL (562)}{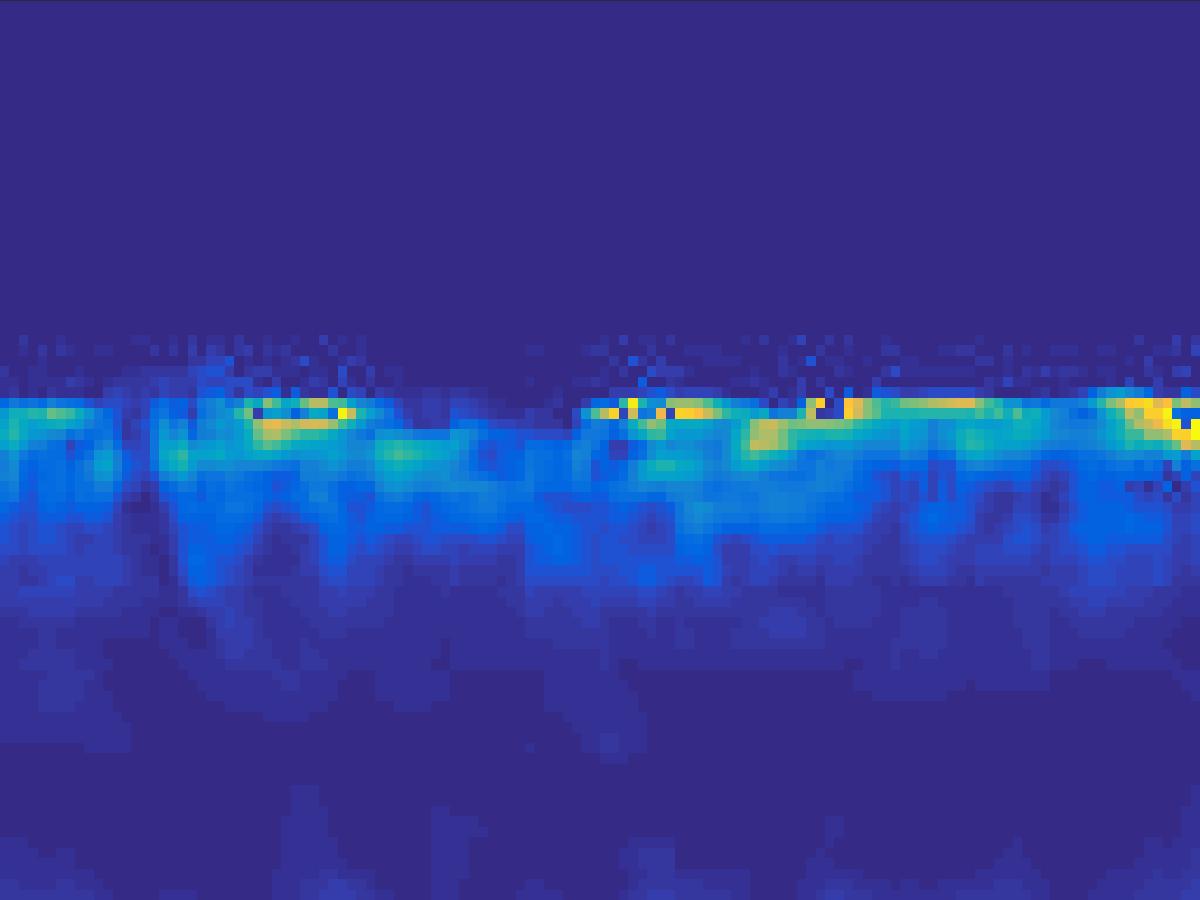}{ncl}  \vspace{-.1in} \\
  \begin{sideways} ~~~~~~~~ Pair-wise difference \end{sideways}&
  \addImgW{Conventional ensemble}{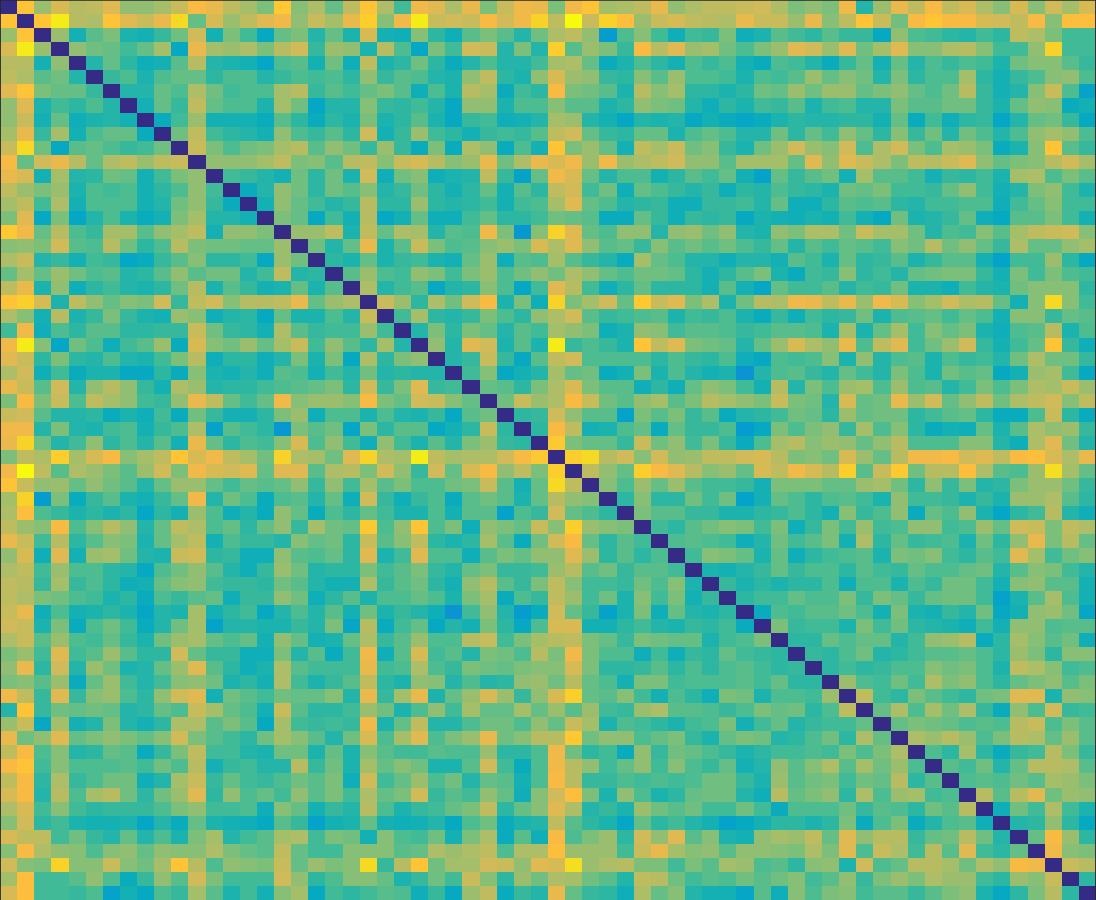}{enseD} &
  \addImgW{DNCL}{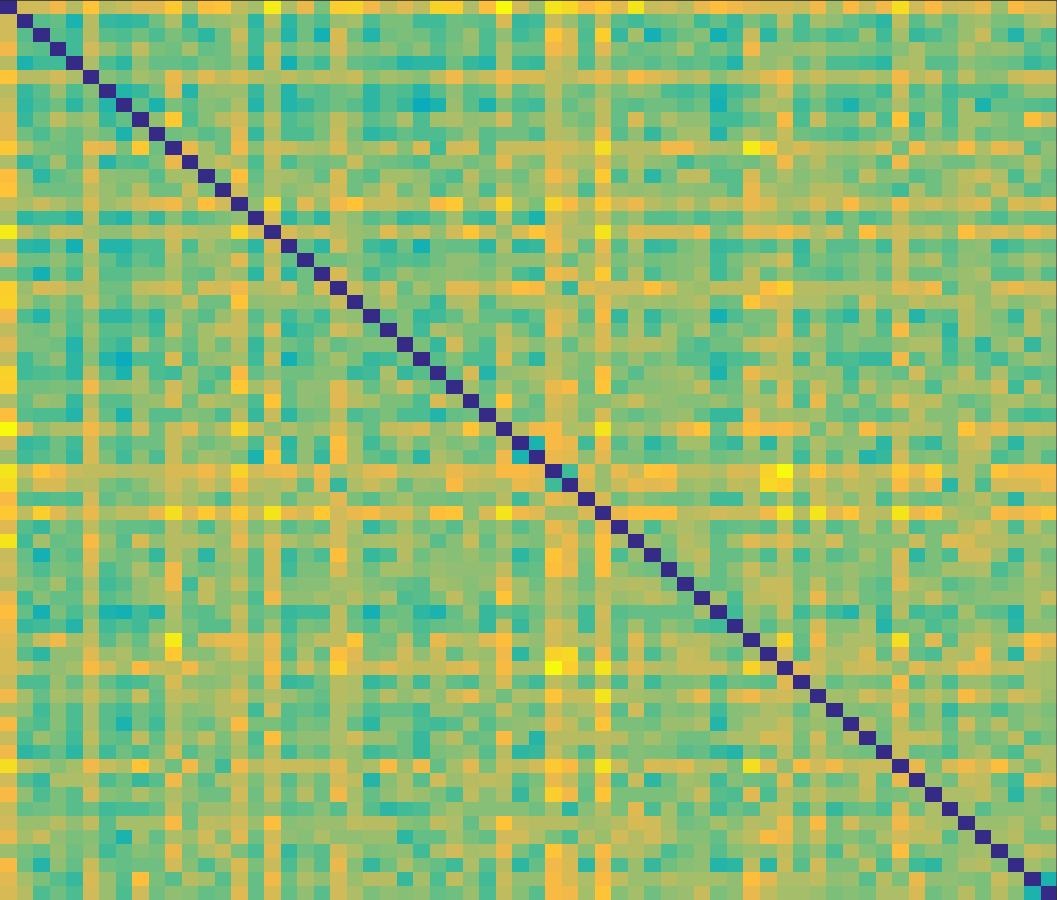}{nclD}  \\
  \end{tabular*}
\caption{Visualization on the diversities with all 64 base models.
  The first row shows  the input image and the ground-truth number of people.
  The second row shows the predicted density map from conventional ensemble and NCL,
  respectively.
  The number in the bracket represents the ground-truth number of people.
  The third row shows the pair-wise Euclidean distance between the predictions
  of individual base models in conventional ensemble and  NCL, respectively.
  It can be seen that the proposed method leads to much diversified base models
  which can yield better overall performances.
}\label{fig:dmap}
\end{figure}

\section{Conclusion}

In this paper, we present a simple yet effective learning strategy for regression.
We pose a typical deep regression network as an ensemble learning problem
and learn a pool of weak regressors using convolutional feature maps.
The main component of this ensemble architecture is the introduction of negative
correlation learning (NCL),
which aims to improve the generalization capability of the ensemble models.
We show the proposed method has sound generalization capability through
managing their intrinsic diversities.
The proposed method is generic and independent of the backbone network
architectures.
Extensive experiments on several challenging tasks including crowd counting,
personality analysis, age estimation and image super-resolution
demonstrate the superiority of the proposed method over other loss functions
and current state-of-the-arts.

\section*{Acknowledgments}
This research was supported by NSFC (NO. 61620106008),
the national youth talent support program, and
Tianjin Natural Science Foundation (17JCJQJC43700, 18ZXZNGX00110).

\ifCLASSOPTIONcaptionsoff \newpage \fi

\bibliographystyle{IEEEtran}
\bibliography{ref}

\newcommand{\AddPhoto}[1]{\includegraphics%
[width=1in,keepaspectratio]{#1}}

\begin{IEEEbiography}[\AddPhoto{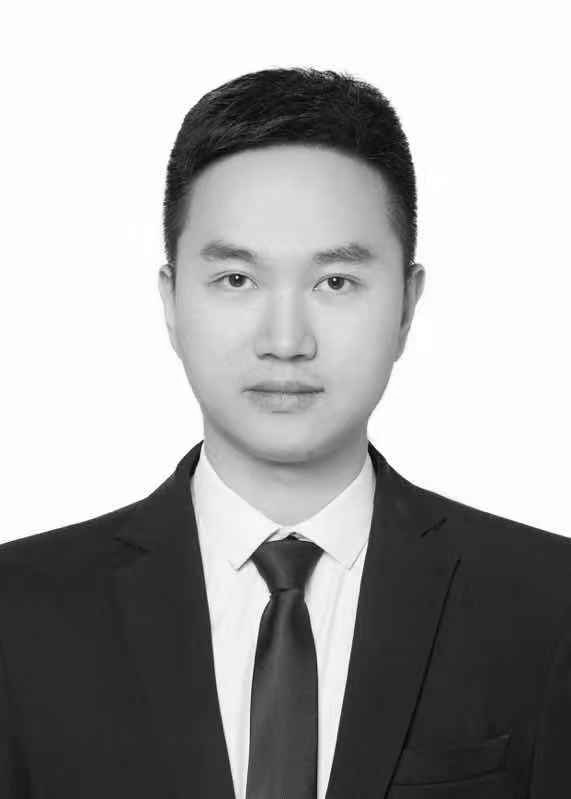}]{Le Zhang}
received the B.Eng degree from University of Electronic
Science and Technology Of China in 2011.
He received his M.Sc and Ph.D.degree
form Nanyang Technological University (NTU) in
2012 and 2016, respectively. Currently, he is a
scientist at Institute for Infocomm Research, Agency
for Science, Technology and Research (A*STAR),
Singapore. He served as TPC member in several conferences such as AAAI, IJCAI. He has served as a Guest Editor for Pattern Recognition and Neurocomputing;
His current research interests include deep learning
and computer vision.
\end{IEEEbiography} \vspace{-0.4in}

\begin{IEEEbiography}[\AddPhoto{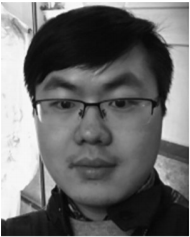}]{Zenglin Shi}
received the Bachelor’s degree engineering in computer science and Master’s degree engineering in computer science from the School of Zhengzhou University, Zhengzhou, China, in 2014 and 2017, respectively. He is currently working toward the Ph.D. degree at the University of Amsterdam, Amsterdam, The Netherlands. His research interests include machine learning, computer vision, and deep learning.
\end{IEEEbiography} \vspace{-0.4in}

\begin{IEEEbiography}[\AddPhoto{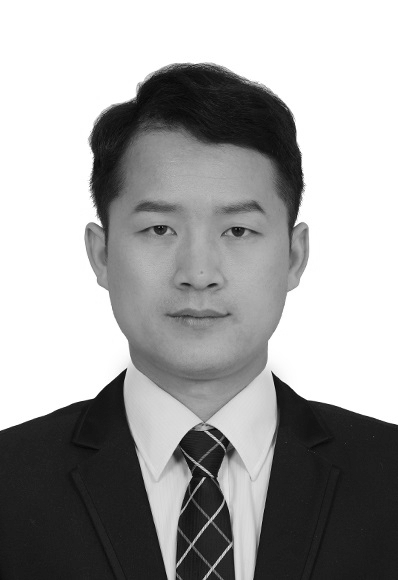}]{Ming-Ming Cheng}
received his PhD degree from Tsinghua University in 2012.
Then he did 2 years research fellow, with Prof. Philip Torr
in Oxford.
He is now a professor at Nankai University, leading the
Media Computing Lab.
His research interests includes computer graphics, computer
vision, and image processing.
He received research awards including ACM China Rising Star Award,
IBM Global SUR Award, and CCF-Intel Young Faculty Researcher Program.
He is on the editorial boards of IEEE TIP.
\end{IEEEbiography} \vspace{-0.4in}

\begin{IEEEbiography}[\AddPhoto{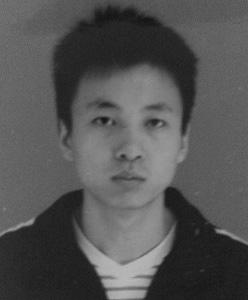}]{Yun Liu}
is a PhD candidate at College of Computer Science,
Nankai University.
He received his bachelor degree from Nankai University in 2016.
His research interests include computer vision and machine learning.
\end{IEEEbiography}\vspace{-0.4in}

\begin{IEEEbiography}[\AddPhoto{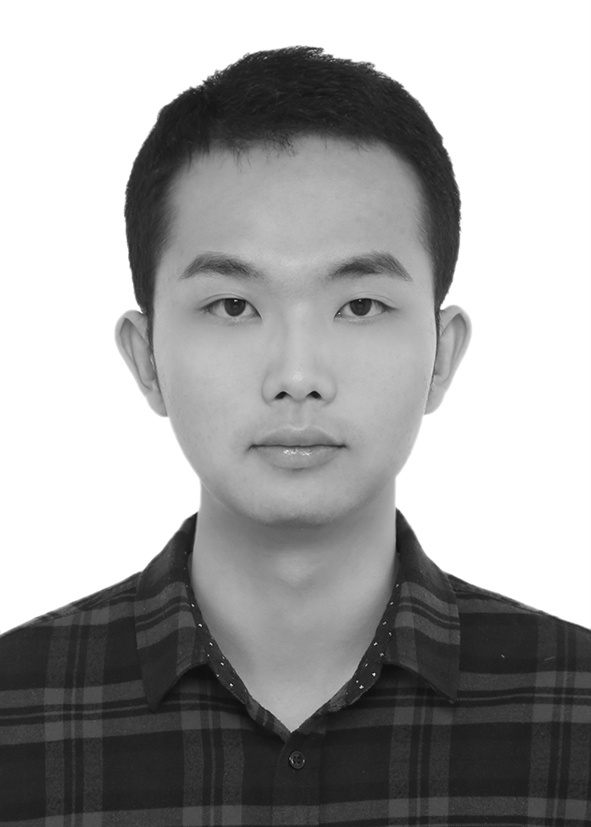}]{Jia-Wang Bian}
is a PhD student at the University of Adelaide and an Associated PhD researcher with the Australian Centre for Robotic Vision (ACRV). He is advised by Prof. Ian Reid and Prof. Chunhua Shen. His research interests lie in the field of computer vision and robotics. Jiawang received his B.Eng. degree from Nankai University, where he was advised by Prof. Ming-Ming Cheng. He was a research assistant at the Singapore University of Technology and Design (SUTD). Jiawang also did a trainee engineer job at the Advanced Digital Sciences Center (ADSC), Huawei Technologies Co., Ltd, and Tusimple.
\end{IEEEbiography} \vspace{-0.4in}

\begin{IEEEbiography}[\AddPhoto{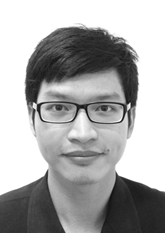}]{Joey Tianyi Zhou}
is a scientist with Institute of High Performance Computing (IHPC), Research Agency for Science, Technology and Research (A*STAR) Singapore. He received his Ph.D. degree in computer science from Nanyang Technological University (NTU), Singapore, in 2015. He was awarded the NIPS 2017 Best Reviewer Award, Best Paper Award at the Beyond Labeler workshop on IJCAI 2016, Best Paper Nomination at ECCV 2016 and Best Poster Honorable Mention at ACML 2012. His research interests include differentiable programming, transfer learning and sparse coding. He served as TPC member in many conferences such as AAAI, IJCAI, ACML and co-organized ACML 2016 Learning on Big Data workshop; He has served as a Guest Editor for IET Image Processing;
\end{IEEEbiography} \vspace{-0.4in}

\begin{IEEEbiography}[\AddPhoto{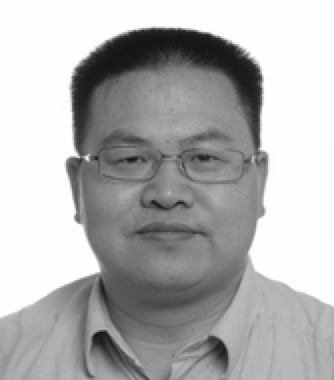}]{Guoyan Zheng}
Guoyan Zheng received the BSc degree and the MSc degree from Southern Medical University, China, in 1992 and 1995 respectively; the PhD degree from the University of Bern, Switzerland, in 2002. He joined the University of Bern in 2003. He did his habilitation and was awarded the title 'Privatdozent' in 2010 and became an Associate Professor in 2015. Starting from 2019, he is a professor in Shanghai Jiao Tong University, China. His research interests include medical image computing, machine learning, computer vision, computer assisted interventions and medical robotics. He is a member of IEEE and the board of directors of the International Society for Medical Image Computing and Computer Assisted Interventions (MICCAI).
\end{IEEEbiography} \vspace{-0.4in}

\begin{IEEEbiography}[\AddPhoto{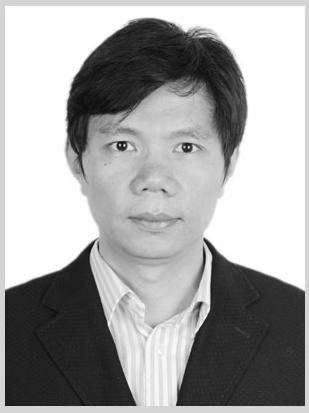}]{Zeng Zeng}
Dr. Zeng Zeng received the Ph.D. degree in electrical and computer engineering from the National University of Singapore, Singapore, in 2005, and the B.S. and M.S. degrees in automatic control from the Huazhong University of Science and Technology, Wuhan, China, in 1997 and 2000, respectively. Currently, he works as Head of Deep Learning for Semiconductor Programme, I2R, A*Star, Singapore. From 2011 to 2014, he worked as a Senior Research Fellow with the National University of Singapore, and he was the Founder of GoGoWise Cloud Education Pte. Ltd., Singapore. From 2005 to 2011, he worked as an Associate Professor in Computer and Communication School, Hunan University, China. From 2008 to 2011, he was a Senior Engineer and Senior Consultant of CSR Zhuzhou Institute Co. Ltd, Hunan, China. In the meanwhile, he was a Senior member of IEC High Speed Train Group and participated in the draft proposals of IEC-61375, IEC-62580, etc. His research interests include deep learning, distributed/parallel computing systems, data stream analysis, multimedia storage systems, wireless sensor networks, onboard fault diagnosis and fault pre-alerting, and controller area networks.
\end{IEEEbiography} \vspace{-0.4in}

\vfill

\end{document}